%% file: main.tex
\documentclass[runningheads]{llncs}

\usepackage{eccv}

\usepackage{eccvabbrv}

\input{preamble}

\usepackage{hyperref}
\usepackage{url}            %
\usepackage{booktabs}       %
\usepackage{amsfonts}       %
\usepackage{nicefrac}       %
\usepackage{microtype}      %
\usepackage{nicematrix}
\usepackage{arydshln}

\usepackage{enumitem}
\usepackage{amsmath}
\usepackage{multirow}
\usepackage{graphicx}
\usepackage{tabularray}
\usepackage{pifont}%
\usepackage{graphicx}
\usepackage{bbm}
\usepackage[accsupp]{axessibility}  %
\usepackage{colortbl}
\usepackage[table]{xcolor}

\newcommand{\methodname}{MegaFlow\xspace}
\title{\methodname: Zero-Shot Large Displacement \\ Optical Flow}

\author{Dingxi Zhang\inst{1} \quad
Fangjinhua Wang\inst{1} \quad
Marc Pollefeys\inst{1,2} \quad
Haofei Xu\inst{1, 3}}

\authorrunning{D. Zhang et al.}

\institute{$^1$ETH Zurich, Switzerland \quad $^2$Microsoft, Switzerland \\ $^3$University of T\"ubingen, T\"ubingen AI Center, Germany \\
\email{zhangdi@ethz.ch, \{fangjinhua.wang, marc.pollefeys, haofei.xu\}@inf.ethz.ch}
}
\begin{document}

\maketitle
\begin{figure}[h]
    \centering
    \vspace{-0.5em}
    \includegraphics[width=\linewidth]{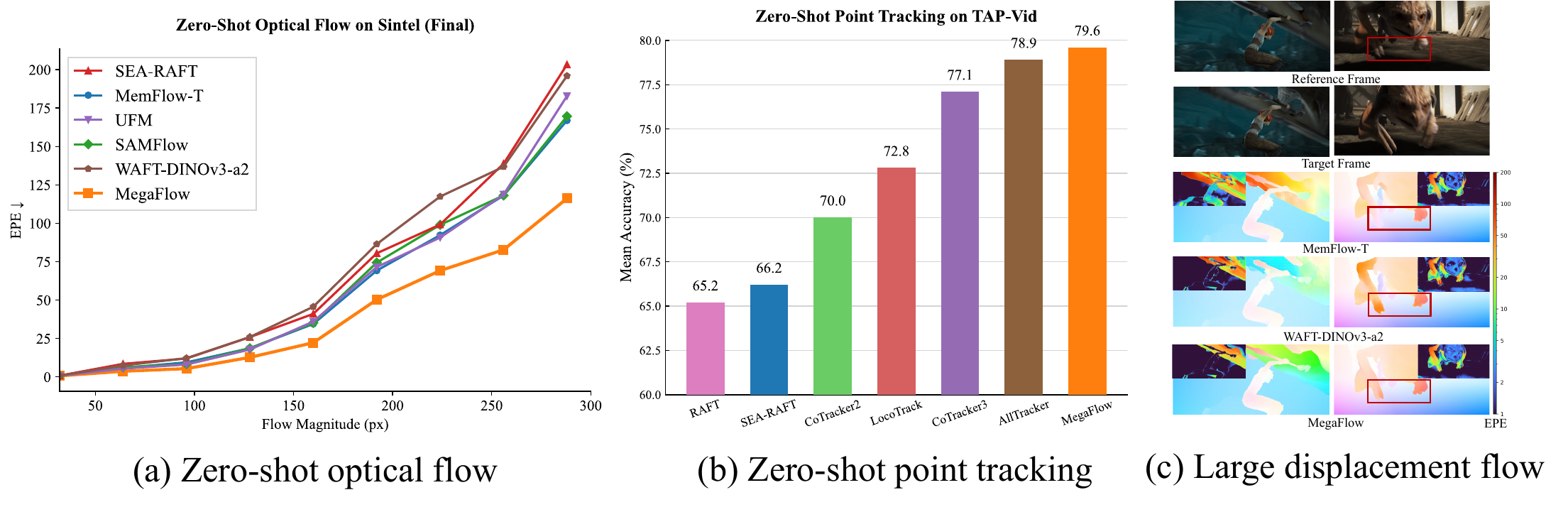}
	
    \captionof{figure}{\textbf{MegaFlow excels at large displacement optical flow and point tracking.} (\textbf{a}) On the Sintel (Final) benchmark, MegaFlow consistently achieves the lowest End-Point Error (EPE), with its advantage widening significantly on large displacements. (\textbf{b}) MegaFlow also demonstrates superior zero-shot point tracking results on TAP-Vid. (\textbf{c}) Visuals and inset error maps further illustrate our state-of-the-art results.}
	\label{fig:teaser}
	\vspace{-2em}
\end{figure}

\input{sec/0_abstract}

\input{sec/1_intro}

\input{sec/2_related}

\input{sec/3_method}

\input{sec/4_exp}

\input{sec/5_conclusion}

\noindent\textbf{Acknowledgments.} This work was supported as part of the Swiss AI Initiative by a grant from the Swiss National Supercomputing Centre (CSCS) under project ID a144 on Alps.
\bibliographystyle{splncs04}
\bibliography{main}
\clearpage
\appendix
\section*{Appendix}

\input{suppl}

\end{document}

%% file: preamble.tex
\usepackage{xcolor} %

%% file: sec/0_abstract.tex
\begin{abstract}

Accurate estimation of large displacement optical flow remains a critical challenge. Existing methods typically rely on iterative local search or/and domain-specific fine-tuning, which severely limits their performance in large displacement and zero-shot generalization scenarios. To overcome this, we introduce \methodname, a simple yet powerful model for zero-shot large displacement optical flow. Rather than relying on highly complex, task-specific architectural designs, \methodname adapts powerful pre-trained vision priors to produce temporally consistent motion fields. In particular, we formulate flow estimation as a global matching problem by leveraging pre-trained global Vision Transformer features, which naturally captures large displacements. This is followed by a few lightweight iterative refinement to further improve the sub-pixel accuracy. Extensive experiments demonstrate that \methodname achieves state-of-the-art zero-shot performance across multiple optical flow benchmarks. Moreover, our model also delivers highly competitive zero-shot performance on long-range point tracking benchmarks, demonstrating its robust transferability and suggesting a unified paradigm for generalizable motion estimation. Project Page: \url{https://kristen-z.github.io/projects/megaflow/}. 

\vspace{-4pt}
\keywords{Optical Flow \and Large Displacement \and Point Tracking }
\vspace{-4pt}

\end{abstract}

%% file: sec/1_intro.tex
\section{Introduction}
\label{sec:intro}

Optical flow estimation is a fundamental problem in computer vision, providing dense pixel-level correspondences that are essential for a wide range of applications, such as autonomous driving~\cite{menze2015object, geiger2012we} and 3D reconstruction~\cite{ma2022multiview, weinzaepfel2023croco}. 

Traditional methods typically formulate optical flow as an optimization problem~\cite{horn1981determining, lucas1981iterative,sun2010secrets}, but struggle with large motions and complex appearance variations. Deep learning dramatically advances this field: early CNN-based approaches~\cite{dosovitskiy2015flownet, ranjan2017optical} demonstrate end-to-end optical flow prediction, while PWC-Net~\cite{sun2018pwc} incorporates pyramid warping and cost volumes to handle large displacements. RAFT~\cite{teed2020raft} further establishes a powerful paradigm using local iterative refinements, inspiring a family of variants emphasizing improved feature aggregation and efficiency~\cite{jiang2021learning_GMA, wang2024sea, wang2025waft}. In parallel, global correspondence matching emerges to complement local refinement~\cite{truong2020glu, xu2022gmflow, xu2023unifying, zhao2022global}, and Transformer-based architectures demonstrate the potential of long-range feature representations for accurate and robust optical flow estimation~\cite{huang2022flowformer, shi2023flowformer++}.

Over the years, the field has advanced considerably, driven by dedicated datasets~\cite{mehl2023spring, kondermann2016hci, menze2015object} and increasingly sophisticated model designs~\cite{huang2022flowformer, shi2023videoflow, sun2025streamflow}. 
Despite these advances, existing methods face two distinct bottlenecks: (1) reliance on task-specific features and domain-specific fine-tuning, which limits out-of-distribution generalization. (2) an architectural vulnerability in resolving large displacements, as iterative local search suffers from severe ambiguities across massive spatial gaps. Consequently, developing a highly accurate and universally generalizable foundation model for optical flow remains an open challenge.

In this paper, we propose \methodname, a general optical flow framework designed for large displacement motion estimation and strong zero-shot generalization. \methodname is inspired by large vision architectures~\cite{wang2025vggt, keetha2025mapanything, wang2025pi3}, whose alternating frame-wise and global-attention Transformer blocks have shown highly effective at modeling cross-view relations. To extend these geometric priors trained on static scenes to dynamic motion estimation, we design a simple yet powerful architecture that builds a globally consistent representation for multi-frame motion estimation. Crucially, this unified design is highly adaptable to diverse vision foundation models and generalizes effectively to different tasks. Instead of directly regressing flow vectors~\cite{zhang2025ufm, weinzaepfel2023croco, lin2025movies}, we first perform global matching to establish accurate initial correspondences. These globally informed predictions are then injected into a lightweight iterative refinement module that uses local correlation to further improve accuracy while preserving fine-grained details. 
\methodname achieves state-of-the-art zero-shot performance on various optical flow and point tracking benchmarks, effectively bridging the gap between powerful static geometric priors and dynamic, large displacement motion estimation.

Our main contributions are summarized as follows:
\vspace{-4pt}
\begin{enumerate}
    \item We introduce \methodname, a simple yet powerful framework for optical flow that significantly improves large displacements and zero-shot generalization.
    
    \item We effectively adapt static pre-trained vision priors to dynamic motion estimation. Our architecture seamlessly integrates global matching with lightweight iterative refinement to capture both large displacements and fine-grained sub-pixel details.
    
    \item Extensive experiments show that \methodname sets a new state of the art for zero-shot optical flow and demonstrates strong cross-task generalization through robust zero-shot point tracking, while also delivering robust performance on in-the-wild video sequences.
\end{enumerate}

%% file: sec/2_related.tex
\section{Related Work}
\label{sec:related}

\noindent\textbf{Optical Flow Estimation.} 
Classical methods typically formulate optical flow as an energy minimization problem~\cite{lucas1981iterative, horn1981determining}, but they usually struggle with complex appearance variations. Early deep learning models enable end-to-end prediction via CNNs and pyramid-based cost volumes~\cite{dosovitskiy2015flownet, ranjan2017optical, sun2018pwc}. Subsequently, RAFT~\cite{teed2020raft} establishes the dominant paradigm of iterative refinement, inspiring many extensions targeting efficiency, aggregation, and memory~\cite{xu2023memory, wang2025waft, jiang2021learning_GMA, wang2024sea, zheng2022dip, morimitsu2025dpflow}. To better capture long-range dependencies, Transformer-based approaches integrate global matching and the attention mechanism~\cite{xu2022gmflow, xu2023unifying, huang2022flowformer, shi2023flowformer++, jiang2021learning_GMA, sui2022craft}, while alternative formulations explore using diffusion models~\cite{saxena2023surprising, luo2024flowdiffuser}. However, two-frame methods inherently lack temporal context. To address this, multi-frame architectures incorporate temporal windows \cite{shi2023videoflow, bargatin2025memfof}, memory buffers ~\cite{dong2024memflow, sun2025streamflow, liu2026arflow} or accumulate correspondences over longer horizons~\cite{wu2023accflow, harley2025alltracker}. Despite these advances, both two-frame and multi-frame frameworks typically train task-specific feature extractors and depend heavily on per-domain fine-tuning, which severely limits their zero-shot generalization across diverse, real-world environments.

\noindent\textbf{Vision Transformers for Optical Flow.} 
Large pre-trained vision Transformers capture exceptionally rich representations that transfer across diverse domains~\cite{wang2025vggt, keetha2025mapanything, oquab2023dinov2, simeoni2025dinov3, depth_anything_v2, wen2025stereo}. For optical flow, existing approaches typically exploit these models by either directly regressing flow via lightweight heads~\cite{weinzaepfel2023croco, saxena2023surprising, zhang2025ufm} or by designing task-specific blocks to process local cost volumes and iterative updates~\cite{huang2022flowformer, shi2023flowformer++, luo2024flowdiffuser, zhou2024samflow, wang2025waft}. While these strategies improve feature robustness, confining powerful static priors to the local search space or unconstrained regression inherently bottlenecks their ability to resolve extreme displacements.

\noindent\textbf{Tracking Any Point.} 
The Tracking Any Point (TAP) task~\cite{doersch2022tap, sand2008particle} evaluates a model's ability to maintain robust, long-range temporal correspondences. Traditionally, optical flow has served as a building block for multi-frame TAP problems~\cite{zheng2023point, ngo2024delta, doersch2024bootstap}. To better handle occlusions and out-of-frame motions, recent methods~\cite{cho2024local, karaev2024cotracker, karaev2024cotracker3} exploit correlations across multiple simultaneous tracks. Concurrently, another emerging line of research demonstrates that leveraging the rich, geometric-aware features from visual foundation models significantly enhances long-term tracking robustness~\cite{aydemir2024can, aydemir2025trackon, aydemir2025trackon2}. Pushing towards unified dense tracking, recent models like AllTracker~\cite{harley2025alltracker} and a concurrent method CoWtracker~\cite{lai2026cowtracker} propose a single-model solution using iterative warping-based refinement. Yet, their reliance on local search paradigms inevitably leads to error accumulation and compromised pixel-level accuracy over time. In contrast, by leveraging generalized global matching and local refinement, \methodname naturally excels at both point tracking and optical flow within a single unified architecture.

\begin{figure*}[t]
    \centering
    \includegraphics[width=\textwidth]{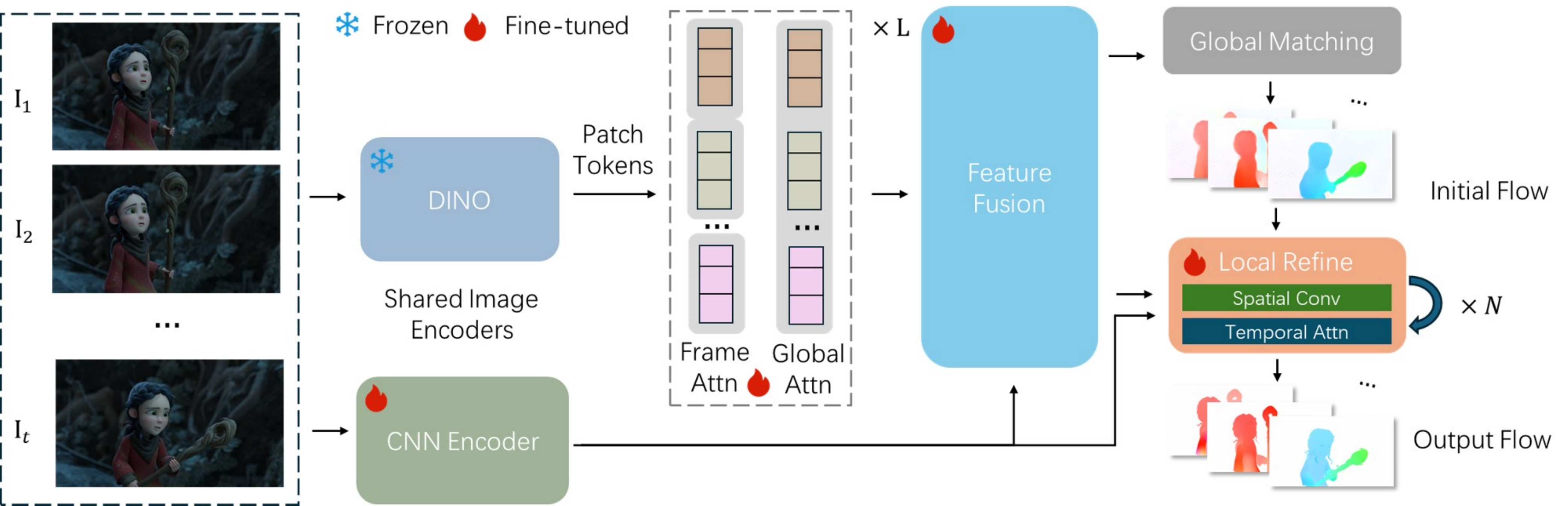} 
    \caption{\textbf{The pipeline of \methodname.} Given an input sequence, a frozen DINO and a trainable CNN extract dense patch tokens and local structural features. Alternating frame and global attention, followed by feature fusion, process these tokens into a globally consistent representation. Pair-wise global matching then computes initial flows. Finally, a recurrent module iteratively refines the initial flows using spatial convolutions and temporal attention for sub-pixel accuracy. Crucially, our design seamlessly processes variable-length inputs without architectural modifications.}
    \label{fig:pipeline}
\end{figure*}

%% file: sec/3_method.tex
\section{Method}
\label{sec:method}
Given a sequence of $T$ video frames $\{I_1, I_2, \dots, I_T\}$, our goal is to estimate the optical flows $\{f_1, \dots, f_{T-1}\} \in \mathbb{R}^{H \times W \times 2}$ between consecutive frames. Specifically, our framework consists of three core components. First, all frames are processed by a shared feature extractor that integrates a trainable local CNN with a frozen vision Transformer backbone. This produces multi-frame feature maps $\{F_1, \dots, F_T\}$ that capture both fine-grained local structures and robust, globally consistent representations. Second, we perform pair-wise global matching between adjacent feature maps to compute an initial flow estimation. Finally, a recurrent refinement module iteratively updates these flow fields using local correlation and temporal attention, yielding accurate and temporally consistent optical flows. An overview of the full architecture is shown in Fig.~\ref{fig:pipeline}.

\subsection{Feature Extraction and Fusion}
Given an input sequence of $T$ frames $\{I_1,\dots,I_T\}$, we first extract per-frame features using DINOv2~\cite{oquab2023dinov2, simeoni2025dinov3}. This produces a set of patch tokens $\{t_i\}_{i=1}^{T}$, which are subsequently processed by a Transformer backbone composed of $L$ alternating {frame-wise} and {global} self-attention layers following VGGT~\cite{wang2025vggt}.

To compensate for the loss of fine spatial details caused by patch tokenization, we introduce a lightweight CNN encoder~\cite{he2016deep} to produce multi-scale feature maps at $1/2$ and $1/4$ resolutions. The $1/4$ resolution features are spatially compressed using pixel unshuffle~\cite{shi2016real} and concatenated with the intermediate Transformer tokens. A DPT-style fusion head~\cite{ranftl2021vision} then aligns and merges the CNN features with the Transformer feature space, producing the fused multi-frame feature maps $\{F_i \in \mathbb{R}^{\frac{H}{7} \times \frac{W}{7} \times D}\}_{i=1}^T$, where $D=128$. These fused features retain strong local details while encoding broad cross-frame context, serving as the input for the global matching and refinement stages.

\subsection{Global Matching}
With the fused feature maps $\{F_i\}_{i=1}^{T}$, we estimate an initial flow between each adjacent pair $(F_i, F_{i+1})$. We formulate correspondence as a global matching problem. For each spatial location $\mathbf{u}$ in $F_i$, we compare it against all locations in $F_{i+1}$ by computing an all-pairs correlation:
\begin{equation}
C_i(\mathbf{u}, \mathbf{v}) = \langle F_i(\mathbf{u}),\, F_{i+1}(\mathbf{v}) \rangle,
\quad C_i \in \mathbb{R}^{\frac{H}{7} \times \frac{W}{7} \times \frac{H}{7} \times \frac{W}{7}},
\label{eq:global-cost}
\end{equation}
where $\langle \cdot, \cdot \rangle$ represents dot product, $\mathbf{u}, \mathbf{v}$ denote pixels. Following~\cite{xu2022gmflow, xu2023unifying}, we apply a softmax normalization over $\mathbf{v}$ to obtain a probability distribution over correspondence candidates:
\begin{equation}
M_i(\mathbf{u}, \mathbf{v}) = \mathrm{softmax}_{\mathbf{v}}\!\left(C_i(\mathbf{u}, \mathbf{v})\right),
\label{eq:matching-dist}
\end{equation}

Let $G$ denote the coordinate grid of $F_{i+1}$. The matched coordinate field is then computed as an expectation over this distribution, and the initial flow is simply the displacement between source and matched coordinates:
\begin{equation}
f^{\text{init}}_i(\mathbf{u}) 
= \sum_{\mathbf{v}} M_i(\mathbf{u}, \mathbf{v}) \cdot G(\mathbf{v}) \;-\; G(\mathbf{u}).
\label{eq:init-flow-grid}
\end{equation}

\subsection{Local Recurrent Refinement}

After obtaining the initial flows $\{f_i^{\text{init}}\}_{i=1}^{T-1}$, we perform iterative refinement using a lightweight recurrent module that jointly leverages local spatial correlations and temporal dependencies.

For each adjacent frame pair $(i, i{+}1)$, the initial flow is bilinearly upsampled to the CNN feature resolution, $\tilde{f}_i^{\text{init}}$, and used to sample the CNN feature of $I_{i+1}$. A local correlation volume is then constructed as:
\begin{equation}
C_i^{\mathrm{local}}(\mathbf{u})
= \left\langle
F_i^{\mathrm{cnn}}(\mathbf{u}),
F_{i+1}^{\mathrm{cnn}}\left(\mathbf{u} + \tilde{f}_i^{\text{init}}(\mathbf{u}) + \Delta\mathbf{u}\right)
\right\rangle,
\label{eq:local-corr}
\end{equation}
where $\mathbf{u}$ denotes a pixel location at $1/4$ resolution and $\Delta\mathbf{u}\in[-r,r]^2$ enumerates local offsets.

The refinement is performed iteratively for $K$ steps, where each update follows: 
\begin{equation}
f_i^{(k+1)} = f_i^{(k)} + \mathcal{R}\left(f_i^{(k)}, C_i^{\mathrm{local}}, F_i^{\mathrm{cnn}}, F_i\right),
\end{equation}
with $\mathcal{R}$ denotes the refinement network and $f_i^{(0)} = \tilde{f}_i^{\text{init}}$.

Specifically, $\mathcal{R}$ consists of two complementary components:
(1) a ConvNeXt-based convolutional branch aggregates local motion evidence from the correlation and CNN features, and
(2) a temporal attention branch correlates features across the sequence.
This hybrid formulation enables the model to capture both fine-grained local motion and long-range temporal coherence, leading to robust flow estimation under occlusions and appearance changes.

Finally, the refinement operates recurrently over all frames, enforcing spatial consistency and temporal smoothness across the sequence.
The design remains agnostic to the number of input frames, allowing \methodname to generalize to variable-length sequences without structural modification.

\subsection{Training Loss}

The model is trained with the following objective for all $T-1$ output flows:
\begin{align}
    \mathcal{L}_{\text{flow}}=\sum_i^{T-1}\left\| f_i^{\text{init}}-\hat{f}_i  \right\|_{\text{smooth}} + \sum_{k=1}^{K}\gamma^{K-k}\sum_i^{T-1}\left\| f_i^k-\hat{f}_i \right\|_1,
\end{align}
where $\hat{f}$ represents ground-truth optical flow, $\left\|\cdot \right\|_{\text{smooth}}$ denotes the smooth $\ell_1$ loss,  $K$ is the iteration number, $\gamma=0.9$ is the weight. We apply exponentially increasing weights~\cite{teed2020raft} to supervise the iteratively refined flow.

\subsection{Extension to Point Tracking}
\label{sec:track_extension}
A key advantage of \methodname is that it seamlessly adapts to dense point tracking without requiring any architectural modifications. Given a video sequence of $T+1$ frames, denoted as $\{I_0, I_1, \dots, I_T\}$, we define $I_0$ as the query frame and the subsequent frames $\{I_1, \dots, I_T\}$ as the targets. Let $\mathbf{x} \in \mathbb{R}^2$ denote a 2D pixel location in image coordinate, and let $\mathcal{P} \subset \mathbb{R}^2$ represent the subset of query locations we aim to track. We operate under the standard assumption that the initial position is anchored at the query coordinate, such that ${p}_0(\mathbf{x}) = \mathbf{x}$.

Our objective is to estimate the corresponding location ${p}_t(\mathbf{x}) \in \mathbb{R}^2$ for every query pixel $\mathbf{x} \in \mathcal{P}$ across all target frames $I_t$, where $t \in \{1, \dots, T\}$. To bridge the task of dense tracking with our multi-frame flow formulation, we parameterize the trajectories using the displacement field. Instead of calculating flow strictly for adjacent frames, our global matching and local refinement stages explicitly compute correspondences $f_{0 \to t}$ between the first frame $I_0$ and each target frame $I_t$. The tracked position of any pixel is directly derived as:
$$ {p}_t(\mathbf{x}) = \mathbf{x} + f_{0 \to t}(\mathbf{x}) $$

In our dense tracking scenario, we define $\mathcal{P}$ to encompass the entire spatial grid of the query frame, effectively tracking all points simultaneously. To optimize the network using sparse point track supervision, we adapt our training objective to directly penalize trajectory errors rather than dense flow fields. The tracking loss is formulated as:
\begin{align}
    \mathcal{L}_{\text{point}} = \sum_{t=1}^{T} \left\| {p}_t^{\text{init}} - \hat{{p}}_t \right\|_{\text{smooth}} + \sum_{k=1}^{K} \gamma^{K-k} \sum_{t=1}^{T} \left\| {p}_t^k - \hat{{p}}_t \right\|_1,
\end{align}
where $\hat{{p}}_t$ represents the ground truth tracked point coordinates at time step $t$. 

To process long sequences, we employ an sliding window strategy with a window size of 8. Specifically, we use the predicted trajectories from the current window as the initialization for the next, sequentially repeating the inference process following \cite{harley2025alltracker}. Unlike dedicated point trackers that depend on explicit visibility heuristics and confidence scores, \methodname propagates tracks entirely through its continuous displacement fields.

%% file: sec/4_exp.tex
\section{Experiments}
\label{sec:exp}
\noindent\textbf{Datasets and Evaluation Protocol.}
We first train on FlyingChairs~\cite{dosovitskiy2015flownet}, TartanAirV1~\cite{tartanair2020iros}, and FlyingThings~\cite{flyingthings}, following standard practices~\cite{teed2020raft, xu2022gmflow, xu2023unifying, wang2024sea, bargatin2025memfof}. Zero-shot evaluation is then conducted on Sintel~\cite{butler2012naturalistic} and KITTI~\cite{geiger2012we} to assess cross-domain generalization. In addition, we train on a mixed dataset comprising FlyingThings~\cite{flyingthings}, HD1K~\cite{kondermann2016hci}, Sintel~\cite{butler2012naturalistic}, and KITTI~\cite{geiger2012we}, and report results on online benchmarks of Sintel, KITTI, and Spring~\cite{mehl2023spring}. Importantly, we do not perform dataset-specific fine-tuning for benchmark submissions. 

\noindent\textbf{Evaluation Metrics.}
We follow standard optical flow evaluation metrics. The primary metric is End-Point Error (EPE), calculated as the average $\ell_2$ distance between the predicted and ground-truth flow. For KITTI, we additionally report F1-all, indicating the percentage of outliers. For the Spring dataset, we include 1-pixel outlier rate (1px), percentage of flow outliers (Fl), and weighted area under the curve (WAUC)~\cite{richter2017playing, geiger2012we, mehl2023spring}. To further analyze performance across motion scales, we report EPE by flow magnitude: $s_{0-10}$, $s_{10-40}$, and $s_{40+}$ correspond to flow magnitudes of $0$–$10$, $10$–$40$, and over $40$ pixels, respectively.

\noindent\textbf{Implementation Details.}
The Transformer backbone consists of $L=24$ layers of alternating global and frame-wise attention following VGGT~\cite{wang2025vggt}. Local features are extracted using the first two blocks of a ResNet~\cite{he2016deep} pretrained on ImageNet~\cite{deng2009imagenet}, producing $1/4$ resolution feature maps. The refinement module comprises two ConvNeXt blocks~\cite{liu2022convnet,wang2024sea} and two temporal attention blocks. The full model contains 936M parameters.

We implement our model with PyTorch~\cite{paszke2019pytorchai} and optimize with AdamW~\cite{loshchilov2017decoupled}. The DINOv2~\cite{oquab2023dinov2} image encoder, Transformer blocks and parts of the feature fusion module are initialized with pre-trained VGGT~\cite{wang2025vggt} weights, and the DINOv2 is kept frozen during training. Training proceeds in 3 stages: (1) 20K iterations on FlyingChairs; (2) 30K iterations on TartanAirV1; (3) 30K iterations on FlyingThings, split into 15K iterations of 2-frame pretraining followed by 15K iterations of multi-frame training; (4) 30K iterations on the mixed dataset. 
The batch size is set to 128 for all stages. The refinement module uses 4 iterations during training and 8 during evaluation. During multi-frame training, we randomly sample between 2 and 6 frames from a random scene, encouraging the model to handle variable temporal spans. By default, inference uses $T=4$ input frames.

The full training process runs on 64 NVIDIA GH200 GPUs over four days. We employ gradient norm clipping with a threshold of 1.0 to ensure stability, and leverage bfloat16 precision and gradient checkpointing to improve memory usage and computational efficiency. All attention layers are accelerated using FlashAttention-3~\cite{shah2024flashattention}. Additional details are provided in the supplementary.

\begin{table}[t]
    \centering
    \caption{\textbf{Zero-shot evaluation on Sintel (train) and KITTI (train)}. Most methods are trained on the FlyingChairs and FlyingThings datasets by default. Note that VideoFlow models \cite{shi2023videoflow} are not trained on FlyingChairs and UFM \cite{zhang2025ufm} are trained on multiple dense correspondence datasets. }
    \begin{tabular}{l c cccc}
    \toprule
     \multirow{2}{*}{Method} & \multirow{2}{*}{Frames} & \multicolumn{2}{c}{Sintel (train)} & \multicolumn{2}{c}{KITTI (train)}\\
    \cmidrule(l{0.5ex}r{0.5ex}){3-4}\cmidrule(l{0.5ex}r{0.5ex}){5-6}
        & & Clean$\downarrow$ & Final$\downarrow$ & Fl-epe$\downarrow$ & Fl-all$\downarrow$ \\ 
    \midrule
        PWC-Net~\cite{sun2018pwc} & \multirow{14}{*}{2} & 2.55 & 3.93 & 10.4 & 33.7\\
        RAFT~\cite{teed2020raft} & & 1.43 & 2.71 & 5.04 & 17.4\\
        GMA~\cite{jiang2021learning_GMA} & & 1.30 & 2.74 & 4.69 & 17.1\\
        DIP~\cite{zheng2022dip} & & 1.30 & 2.82 & 4.29 & 13.7\\
        RPKNet~\cite{morimitsu2024recurrent} & & 1.12 & 2.45 & - & 13.0\\
        GMFlow~\cite{xu2022gmflow} & & 1.08 & 2.48 & 11.20 & 28.7 \\
        FlowFormer~\cite{huang2022flowformer} & & 1.01 & 2.40 & 4.09 & 14.7\\
        Flowformer++~\cite{shi2023flowformer++} & & 0.90 & 2.30 & 3.93 & 14.2\\ 
        SEA-RAFT (L)~\cite{wang2024sea} & & 1.19 & 4.11 & 3.62 & 12.9\\
        AnyFlow~\cite{jung2023anyflow} & & 1.10 & 2.52 & 3.76 & 12.4 \\
        SAMFlow~\cite{zhou2024samflow} & & 0.87 & 2.11 & 3.44 & 12.3 \\
        FlowDiffuser~\cite{luo2024flowdiffuser} & & \underline{0.86} & 2.19 & 3.61 & 11.8 \\
        WAFT-DINOv3-a2~\cite{wang2025waft} & & 1.28 & 2.56 & 3.49 & 12.9 \\
        UFM \cite{zhang2025ufm} & & 1.15 & \underline{2.01} & \textbf{2.96} & 11.0 \\
        \methodname & & 0.89 & 2.07 & \underline{3.00} & \textbf{10.6} \\
        \midrule
        VideoFlow-BOF$^{*}$~\cite{shi2023videoflow} & 3 & 1.03 & 2.19 & 3.96 & 15.3 \\
        VideoFlow-MOF$^{*}$~\cite{shi2023videoflow} & 5 & 1.18 & 2.56 & 3.89 & 14.2 \\
        StreamFlow~\cite{sun2025streamflow} & 4 & 0.87 & 2.11 & 3.85 & 12.6 \\
        MemFlow~\cite{dong2024memflow} & 3 & 0.93 & 2.08 & 3.88 & 13.7 \\
        MemFlow-T~\cite{dong2024memflow} & 3 & \textbf{0.85} & 2.06 & 3.38 & 12.8 \\
        \methodname & 4 & \textbf{0.85} & \textbf{1.83} & 3.20 & \underline{10.7} \\
    \bottomrule
    \end{tabular}
    
    \label{tab:zero-shot}
\end{table}

\begin{table}[t]
    \centering
    \caption{\textbf{Optical flow estimation across different motion magnitudes}. \methodname significantly reduces EPE on extreme large displacements ($s_{40+}$).
    }
    \label{tab:magnitude_metrics}
    
    \renewcommand{\arraystretch}{1.15} 
    
    \vspace{-4pt}
    \begin{tabular}{l ccc ccc}
    \toprule
    \multirow{2}{*}{Method} & \multicolumn{3}{c}{Sintel (Clean) $\downarrow$} & \multicolumn{3}{c}{Sintel (Final) $\downarrow$} \\
    \cmidrule(lr){2-4} \cmidrule(lr){5-7} 
    & $s_{40+}$ & $s_{10-40}$ & $s_{0-10}$ & $s_{40+}$ & $s_{10-40}$ & $s_{0-10}$ \\
    \midrule
    SEA-RAFT \cite{wang2024sea}    & 8.286 & 1.343 & 0.261 & 26.878 & 4.259 & 0.547 \\
    MemFlow-T \cite{dong2024memflow}    & 5.239 & \underline{0.980} & \textbf{0.211} & 13.670 & 2.224 & \underline{0.371} \\
    SAMFlow  \cite{zhou2024samflow}   & \underline{5.117} & 1.029 & \underline{0.215} & 13.575 & 2.429 & 0.396 \\
    UFM  \cite{zhang2025ufm}    & 6.836 & 1.209 & 0.259 & \underline{12.963} & 2.129 & 0.399 \\
    WAFT-DINOv3-a2 \cite{wang2025waft} & 8.870 & 1.218 & 0.217 & 13.192 & \underline{1.966} & \textbf{0.324} \\
    \methodname & \textbf{4.729} & \textbf{0.909} & 0.314 & \textbf{11.175} & \textbf{1.941} & 0.480 \\
    \bottomrule
    \end{tabular}
\end{table}

\vspace{-2pt}
\subsection{Zero-Shot Generalization}
\label{sec:zero-shot}
We evaluate the zero-shot performance of \methodname on the Sintel and KITTI training sets after the third stage training. 
As shown in Tab.~\ref{tab:zero-shot}, \methodname establishes a new state-of-the-art, outperforming both two-frame and multi-frame architectures. On Sintel, while maintaining a highly competitive Clean EPE of 0.85, \methodname significantly advances the Final EPE to an unprecedented 1.83. Because the Sintel Final pass introduces severe motion blur, atmospheric effects, and complex occlusions, this substantial margin demonstrates the exceptional robustness of \methodname against extreme appearance changes. 

Furthermore, on KITTI benchmark, \methodname achieves the best overall Fl-all error of 10.7 and a highly competitive Fl-epe of 3.00. Notably, \methodname overall outperforms recent models with similar vision priors like UFM~\cite{zhang2025ufm}, despite UFM was trained on significantly more diverse dense correspondence datasets. Overall, the zero-shot evaluation confirms that our approach balances precision and generalization, effectively handling diverse motion dynamics across both synthetic and real-world benchmarks.

\begin{table}[t]
    \centering
    \caption{\textbf{Zero-shot and fine-tuned point tracking comparison} on TAP-Vid \cite{doersch2022tap} benchmarks. 
    We evaluate $\delta_{avg}$ using an input resolution of $384\times512$. Note that the optical flow models are trained solely on mixed optical flow datasets, while point trackers are trained on tracking datasets. Despite this, our zero-shot flow model achieves competitive performance comparable to dedicated trackers. After fine-tuning (`Flow $\rightarrow$ Kubric'), \methodname establishes state-of-the-art results.
    }
    \label{tab:track-comp}
    \vspace{-4pt}

    \begin{tabular*}{\textwidth}{@{\extracolsep{\fill}} l l c c c c @{}}
    \toprule
    Method & Training & DAVIS $\uparrow$ & Kinetics $\uparrow$ & RGB-Stacking $\uparrow$ & Mean $\uparrow$ \\
    \midrule
    PIPs++~\cite{zheng2023point} & PointOdyssey & 62.5 & 64.2 & 70.4 & 65.7 \\
    LocoTrack~\cite{cho2024local} & Kubric & 68.0 & 70.0 & 80.3 & 72.8 \\
    BootsTAPIR \cite{doersch2024bootstap} & Kubric+15M & 67.9 & 70.6 & 81.0 & 73.1 \\
    CoTracker2~\cite{karaev2024cotracker} & Kubric & 70.9 & 65.8 & 73.4 & 70.0 \\
    CoTracker3-Kub~\cite{karaev2024cotracker3} & Kubric & 77.4 & 70.6 & 83.4 & 77.1 \\
    CoTracker3~\cite{karaev2024cotracker3} & Kubric+15K & 77.1 & 71.8 & 84.2 & 77.7 \\
    AllTracker-Kub~\cite{harley2025alltracker} & Kubric & 75.2 & 71.3 & 90.1 & 78.9 \\
    AllTracker~\cite{harley2025alltracker} & Kubric+mix & 76.3 & \textbf{72.3} & 90.0 & 79.5 \\
    \methodname & Flow $\rightarrow$ Kubric & \textbf{77.6} & 70.2 & \textbf{91.1} & \textbf{79.6} \\
    \midrule
    RAFT~\cite{teed2020raft} & Flow mix& 48.5 & 64.3 & 82.8 & 65.2 \\
    SEA-RAFT~\cite{wang2024sea} & Flow mix& 48.7 & 64.3 & 85.7 & 66.2 \\
    AccFlow~\cite{wu2023accflow} & Flow mix& 23.5 & 38.8 & 63.2 & 41.8 \\
    MemFlow-T~\cite{dong2024memflow} & Flow mix & 61.7 & 64.9 & 85.2 & 70.6 \\
    WAFT-DINOv3-a2~\cite{wang2025waft} & Flow mix & 53.9 & 61.0 & 84.0 & 66.3 \\
    \methodname & Flow mix & \textbf{65.6} & \textbf{65.5} & \textbf{89.6} & \textbf{73.6} \\
    \bottomrule
    \end{tabular*}
    \vspace{-2mm}
\end{table}

\begin{figure}[t]
    \centering
    \includegraphics[width=\linewidth]{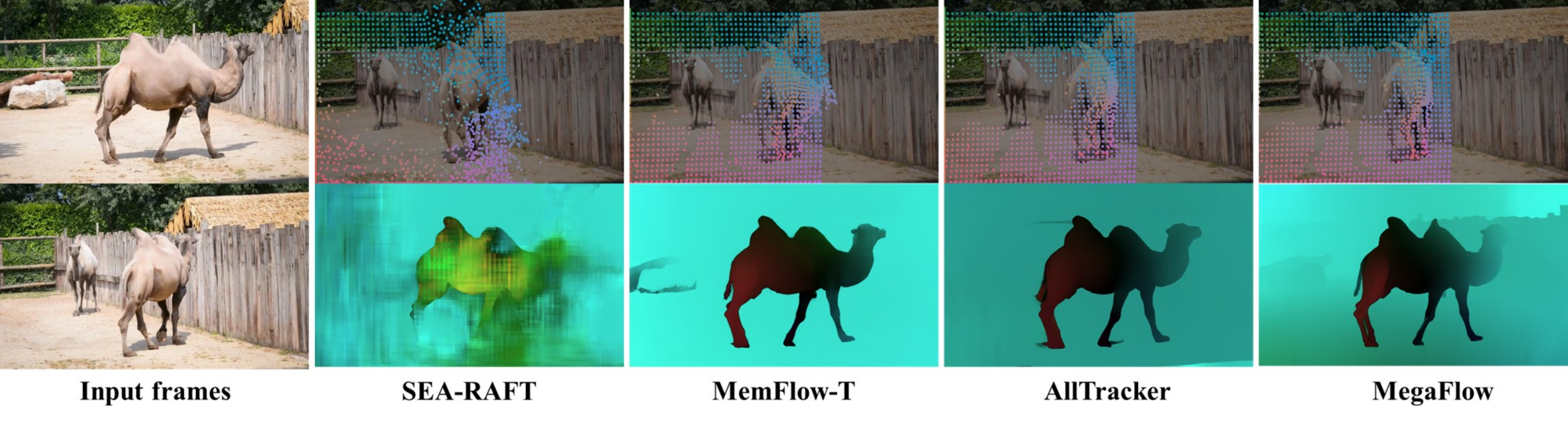}
  
    \caption{\textbf{Qualitative comparison of long-range point tracking.}  Visualization of SEA-RAFT \cite{wang2024sea}, MemFlow \cite{dong2024memflow} and AllTracker \cite{harley2025alltracker} and our method on the DAVIS benchmark. The first column shows the input frames (spanning 90 frames). The top row visualizes long-range dense point tracking, while the bottom row shows the corresponding optical flow between the first and last frame. Our method produces more accurate and temporally consistent tracks and flow estimates over very long sequences.}
    \vspace{-8pt}
    \label{fig:viz_tracking}
\end{figure}

\subsection{Large Displacement Optical Flow}
\label{sec:large_flow}

To explicitly assess robustness against extreme motion, we evaluate \methodname in a zero-shot setting across different flow magnitude intervals on the Sintel dataset (Tab.~\ref{tab:magnitude_metrics}). Following the evaluation protocol from Sec.~\ref{sec:zero-shot}, we compare our approach against state-of-the-art two-frame methods~\cite{wang2024sea, zhang2025ufm, wang2025waft, zhou2024samflow} and the multi-frame architecture MemFlow-T~\cite{dong2024memflow}.

As illustrated by the error curves in Fig.~\ref{fig:teaser}(a), the End-Point Error (EPE) of baseline methods escalates rapidly as motion magnitude increases. The quantitative breakdown in Tab.~\ref{tab:magnitude_metrics} further confirms this vulnerability: although recent iterative-based baselines perform competitively on minor displacements ($s_{0-10}$), their accuracy degrades severely in the extreme $s_{40+}$ regime. While multi-frame models like MemFlow-T attempt to mitigate this by accumulating sequential memory, they remain inherently bottlenecked by localized search paradigms. In contrast, \methodname effectively flattens this error curve and establishes a significant margin of improvement on extreme displacements, driving the $s_{40+}$ error down to a remarkable 4.729 on Sintel (Clean) and 11.175 on Sintel (Final). 

\begin{figure}[htbp]
    \centering
    \includegraphics[width=\linewidth]{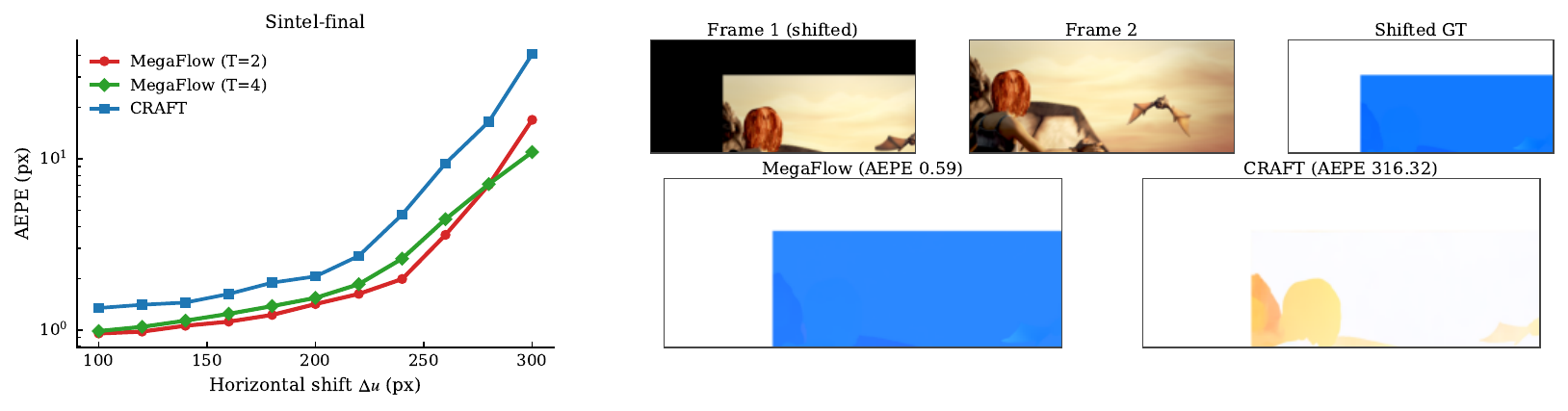}
    \caption{\textbf{Robustness to massive synthetic shifts.} Quantitative error curves (left) and qualitative visual results (right) under the CRAFT shift protocol. \methodname maintains stable predictions at extreme displacements where the baseline catastrophically fails.}
    \label{fig:craft}
\end{figure}

To verify that this capability extends beyond naturally occurring dataset statistics, we further evaluate our model against massive synthetic displacements using the shift protocol introduced by CRAFT \cite{sui2022craft}. We applied this protocol to the Sintel (train) set, evaluating the Average End-Point Error (AEPE) on valid regions for shifts of $\Delta u \in \{100, \dots, 300\}$ and $\Delta v = \Delta u/2$. As shown in Fig.~\ref{fig:craft}, \methodname consistently outperforms CRAFT across all shift magnitudes. Furthermore, qualitative results in the right demonstrate that for these massive shifts, \methodname successfully recovers the ground truth, whereas CRAFT catastrophically reverses the predicted direction. This confirms that our unified architecture generalizes robustly to artificial large motions just as effectively as natural ones.

\subsection{Long-Range Point Tracking}
\vspace{-4pt}
\label{sec:long_tracking}

To assess representation robustness on long-range correspondences, we evaluate \methodname on TAP-Vid~\cite{doersch2022tap} benchmarks (including Kinetics, DAVIS, RGB-Stacking datasets) \emph{without architectural modifications}. For zero-shot evaluation, we applying the procedure in Sec. \ref{sec:track_extension} to all flow-based baselines. To establish our architecture's full tracking potential, we evaluate a variant fine-tuned on Kubric~\cite{greff2022kubric} for 20K iterations after trained on mixed flow dataset. Notably, our sliding window approach processes up to 600-frame sequences on a single GH200 GPU.

Following standard protocols, we evaluate all models at $384\times512$ resolution using the $\delta_{\text{avg}}$ metric~\cite{harley2025alltracker, karaev2024cotracker, karaev2024cotracker3}, which averages $\delta_k=100\cdot\mathbf{1}[\|p-\hat{p}\|_2<k]$ across $k \in \{1,2,4,8,16\}$. Baselines include state-of-the-art trackers~\cite{zheng2023point, cho2024local, karaev2024cotracker, karaev2024cotracker3, doersch2024bootstap} and flow-based methods~\cite{teed2020raft, wang2024sea, wu2023accflow, wang2025waft, dong2024memflow}.

As shown in Tab.~\ref{tab:track-comp} and Fig.~\ref{fig:teaser}(b), \methodname exhibits exceptional cross-task transferability. In the strictly zero-shot setting, it achieves a 73.6\% average accuracy, significantly outperforming all flow baselines and surpassing dedicated trackers like PIPs++, LocoTrack, and CoTracker2. On RGB-Stacking, our zero-shot model exceeds nearly all task-specific architectures. Fine-tuned solely on Kubric (`Flow $\rightarrow$ Kubric'), \methodname establishes a new state-of-the-art average of 79.6\%, decisively outperforming CoTracker3 and AllTracker despite their exposure to more extensive tracking datasets.

Qualitatively (Fig.~\ref{fig:viz_tracking}), \methodname consistently yields accurate, coherent dense tracks. Unlike local search methods (e.g., SEA-RAFT) that produce unstable trajectories on long-range motions, or MemFlow which suffers from boundary artifacts around articulated limbs, our global matching architecture remains robust. Furthermore, while AllTracker achieves reasonable coherence on dense grids, it lacks pixel-level accuracy. Conversely, \methodname maintains sharp local structures across extended sequences, successfully bridging long-range stability with high-fidelity flow estimation.

Ultimately, our unified flow formulation generalizes exceptionally well to long-range tracking. The zero-shot performance proves that foundation vision priors and global matching transfer effectively without tracking-specific supervision, while fine-tuning confirms \methodname as a highly capable architecture for generalized dense motion estimation. Additional in-the-wild visual results are provided in the supplementary material.

\begin{table*}[t!]
\centering
\caption{\textbf{Benchmark comparison of optical flow methods on the Spring test set.} \methodname achieves state-of-the-art zero-shot performance and remains highly competitive without any dataset-specific adaptation
}
\vspace{-4pt}
\begin{tabular}{l l c c c c c}
\toprule
& Method & Frames & EPE $\downarrow$ & Fl $\downarrow$ & WAUC $\uparrow$ & 1px $\downarrow$ \\
\midrule
\multirow{10}{*}{w/o Fine-tune}
 & PWC-Net~\cite{sun2018pwc} & 2 & 2.288 & 4.889 & 45.670 & 82.265 \\
 & RAFT~\cite{teed2020raft} & 2 & 1.476 & 3.198 & 90.920 & 6.790 \\
 & GMA~\cite{jiang2021learning_GMA} & 2 & 0.914 & 3.079 & 90.722 & 7.074 \\
 & FlowFormer~\cite{huang2022flowformer} & 2 & 0.723 & 2.384 & 91.679 & 6.510 \\
 & RPKNet~\cite{morimitsu2024recurrent} & 2 & 0.657 & 1.756 & 92.638 & 4.809 \\
 & MemFlow~\cite{dong2024memflow} & 3 & 0.627 & 2.114 & 92.253 & 5.759 \\
 & StreamFlow~\cite{sun2025streamflow} & 4 & 0.606 & 1.856 & 93.253 & 5.215 \\
 & \methodname & 2 & 0.398 & 1.417 & 92.595 & \textbf{3.949} \\
 & \methodname & 4 & \textbf{0.349} & \textbf{1.261} & \textbf{93.738} & 4.521 \\
\midrule
\multirow{5}{*}{w/ Fine-tune}
 & CrocoFlow~\cite{weinzaepfel2023croco} & 2 & 0.498 & 1.508 & 93.660 & 4.565 \\
 & SEA-RAFT~(M)~\cite{wang2024sea} & 2 & 0.363 & 1.347 & 94.534 & 3.686 \\
 & MemFlow~\cite{dong2024memflow} & 3 & 0.471 & 1.416 & 93.855 & 4.482 \\
 & StreamFlow~\cite{sun2025streamflow} & 4 & 0.467 & 1.424 & 94.404 & 4.152 \\
 & WAFT-DINOv3-a2 \cite{wang2025waft} & 2 & \textbf{0.325} & \textbf{1.246} & \textbf{95.051} & \textbf{3.289} \\
\bottomrule
\end{tabular}
\vspace{-5pt}
\label{tab:spring-comparison}
\end{table*}
\begin{figure*}[htbp]
    \centering
    \includegraphics[width=\linewidth]{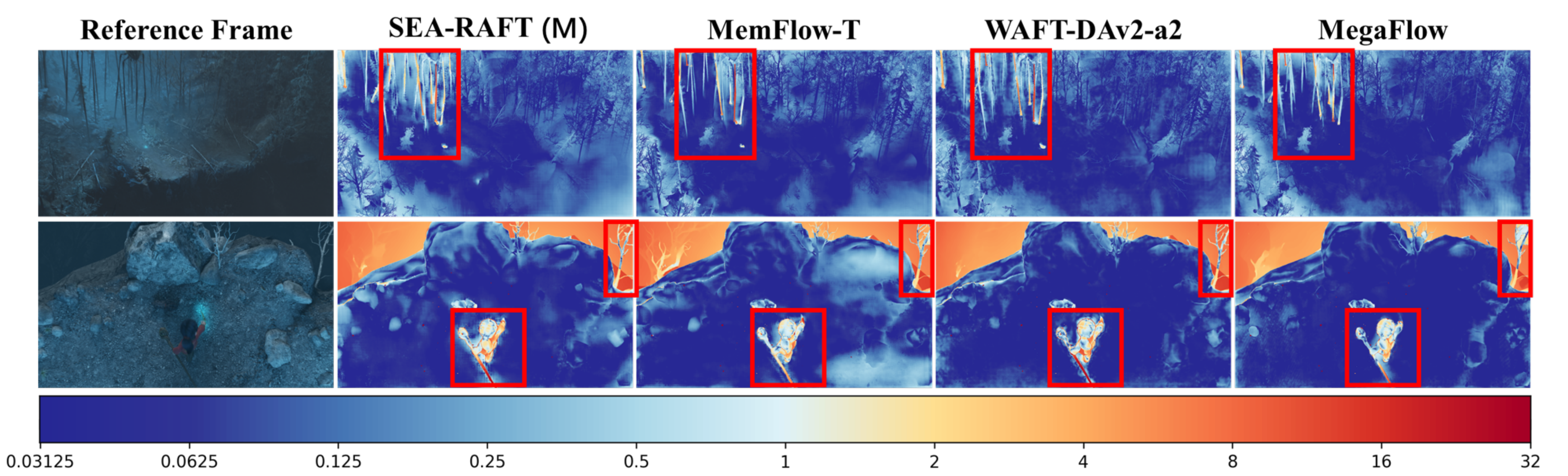}
  
    \caption{\textbf{Qualitative comparison of optical flow.} Visualization of SEA-RAFT \cite{wang2024sea}, MemFlow \cite{dong2024memflow}, WAFT-DAv2-a2 \cite{wang2025waft}, and our method on the Spring benchmark. The colorbar indicates endpoint error. Our approach outperforms prior methods, demonstrating that our method generalizes well to Full HD resolution while preserving both local and global motion details. 
    }

    \label{fig:viz_flow}
\end{figure*}
\begin{table}[t!]
\centering
\caption{\textbf{Benchmark results on Sintel (test) and KITTI (test)}. Methods are separated into two-frame and multi-frame models. We explicitly report results on large-displacement motions ($s_{40+}$) for the Sintel benchmark. 
}
\label{tab:benchmark}
\vspace{-6pt}

\begin{tabular}{l cc cc c}
\toprule
\multirow{2}{*}{Method} & \multicolumn{2}{c}{Sintel (Clean) $\downarrow$} & \multicolumn{2}{c}{Sintel (Final)$\downarrow$} & KITTI $\downarrow$ \\
\cmidrule(lr){2-3} \cmidrule(lr){4-5} \cmidrule(lr){6-6}
 & EPE & $s_{40+}$ & EPE & $s_{40+}$ & Fl-all \\
\midrule
PWC-Net~\cite{sun2018pwc} & 3.86 & 25.29 & 5.04 & 31.07 & 9.60 \\
RAFT~\cite{teed2020raft}& 1.61 & 9.28 & 2.86 & 16.37 & 5.10 \\
GMA~\cite{jiang2021learning_GMA} & 1.39 & 7.60 & 2.47 & 13.50 & 5.15\\
SEA-RAFT~(L)~\cite{wang2024sea} & 1.31 & 7.90 & 2.60 & 15.71 & 4.30 \\
FlowFormer~\cite{huang2022flowformer} & 1.16 & 6.44 & 2.09 & 11.67 & 4.68 \\
RPKNet~\cite{morimitsu2024recurrent} & 1.32 & 7.28 & 2.66 & 16.00 & 4.64 \\
CrocoFlow~\cite{weinzaepfel2023croco} & 1.09 & 6.30 & 2.44 & 15.21 & 3.64 \\
DDVM~\cite{saxena2023surprising} & 1.75 & 12.22 & 2.48 & 16.55 & \textbf{3.26} \\
AnyFlow~\cite{jung2023anyflow} & 1.21 & 7.32 & 2.44 & 14.20 & 4.41 \\
SAMFlow~\cite{zhou2024samflow} & 1.00 & 5.25 & 2.08 & 11.28 & 4.49 \\
FlowDiffuser~\cite{luo2024flowdiffuser} & 1.02 & 5.57 & 2.03 & 10.93 & 4.17 \\
DPFlow~\cite{morimitsu2025dpflow} & 1.05 & 5.84 & 1.98 & 11.39 & 3.56 \\
WAFT-DINOv3-a2~\cite{wang2025waft} & \underline{0.95} & 5.52 & 2.02 & 12.49 & \underline{3.56} \\
VideoFlow-BOF~\cite{shi2023videoflow} & 1.01 & 5.61 & \underline{1.71} & \underline{9.42} & 4.44 \\
VideoFlow-MOF~\cite{shi2023videoflow} & 0.99 & 5.48 & \textbf{1.65} & \textbf{8.80} & 3.65 \\
StreamFlow~\cite{sun2025streamflow} & 1.04 & 6.00 & 1.87 & 10.68 & 4.24 \\
MemFlow-T~\cite{dong2024memflow} & 1.08 & 6.02 & 1.83 & 9.83 & 3.88 \\
\methodname & \textbf{0.91} & \textbf{4.84} & 2.43 & 15.50 & 3.94 \\
\bottomrule
\end{tabular}
\vspace{-6pt}
\end{table}

\vspace{-8pt}
\subsection{Benchmark Results}
\vspace{-4pt}

\noindent\textbf{Spring.} Without fine-tuning, \methodname achieves state-of-the-art zero-shot performance on the Spring benchmark, obtaining the lowest EPE and Fl-all scores in Tab.~\ref{tab:spring-comparison}. Notably, it surpasses most models specifically fine-tuned on high-resolution ($1080\times 1920$) data, demonstrating robust generalization despite being trained solely on standard-resolution datasets ($432\times 960$). This scalability stems from our global initialization, which provides a strong cross-frame prior, and a hybrid refinement module that effectively integrates spatial and cross-frame cues across varying resolutions. While WAFT~\cite{wang2025waft} exhibits marginally higher accuracy through high-resolution supervision, \methodname remains highly competitive without any dataset-specific adaptation, underscoring its inherent architectural scalability. Qualitative results (Fig.~\ref{fig:viz_flow}) show that \methodname produces sharper motion boundaries and preserves finer details than prior methods, particularly for thin structures such as tree branches and slender objects.

\noindent\textbf{Sintel and KITTI.} 
As shown in Tab.~\ref{tab:benchmark}, \methodname achieves competitive performance using a \textit{single, unified model} without dataset-specific fine-tuning. On Sintel (Clean), we reach a state-of-the-art EPE of 0.91 and exhibit superior robustness on extreme displacements ($s_{40+}$) with an EPE of 4.84. While many specialized methods rely on per-benchmark adaptation, our approach prioritizes the general-purpose motion representations inherent in foundation models. Consistent with~\cite{saxena2023surprising, wang2025waft}, we observe that the `Ambush 1' sequence in Sintel (Final) remains a significant outlier. As detailed in the supplementary material, excluding this anomalous sequence reveals that \methodname outperforms recent architectures such as WAFT and MemFlow, while achieving results fully comparable to the VideoFlow framework. This confirms our model's superior capability in handling complex, occluded motion. On KITTI, \methodname yields a competitive Fl-all. This performance is primarily influenced by the fixed-patch tokenization of the Transformer backbone, which is sensitive to KITTI's specific resolution and aspect ratio. Unlike baselines employing exhaustive multi-scale inference or specialized padding, we maintain a zero-shot-style evaluation to verify intrinsic transferability. In this context, \methodname remains on par with other architectures with pretrained prior~\cite{weinzaepfel2023croco, wang2025waft, zhou2024samflow}, effectively balancing high precision with broad generalization. See supplementary for more qualitative benchmark results.

\begin{table}[htbp]
    \centering
    \caption{\textbf{Ablation of Pre-trained Priors and Architecture.} Evaluated zero-shot using 2 input frames. The inference latency is measured on an RTX 3090 at $540\times 960$.}
    \label{tab:ablation_arch}
    \vspace{-6pt}
    
    \resizebox{\linewidth}{!}{%
    \begin{tabular}{l l c c c c c c c}
    \toprule
    & Variant & Param( M) & Latency (ms) & Memory (GB) & S-Clean $\downarrow$ & S-Final $\downarrow$ & K-epe $\downarrow$ & K-all $\downarrow$ \\
    \midrule
    \multirow{3}{*}{{w/o Pre-train}}
    & 6 layers & 256 & 148.4 & 2.93 & 2.05 & 2.46 & 9.71 & 17.9 \\
    & 12 layers & 483 & 209.5 & 4.26 & 2.13 & 3.31 & 11.70 & 19.8 \\
    & Conv embedder & 632 & 262.1 & 4.95 & 1.05 & 2.38 & 4.89 & 15.4 \\
    \midrule
    \multirow{5}{*}{{w/ Pre-train}}
    & 6 layers & 256 & 148.4 & 2.93 & 1.22 & 2.30 & 7.14 & 16.1 \\
    & 12 layers & 483 & 209.5 & 4.26 & 1.11 & 2.23 & 6.76 & 14.4 \\
    & Freeze transformer & 936 & 327.9 & 6.08 & 1.22 & 3.17 & 5.13 & 14.7 \\
    & w/o feat fusion & 935 & 321.1 & 5.81 & 1.02 & 2.11 & 4.71 & 13.9 \\
    \midrule
    & Full Model & 936 & 327.9 & 6.08 & \textbf{0.89} & \textbf{2.08} & \textbf{3.00} & \textbf{10.9} \\
    \bottomrule
    \end{tabular}
    }
    \vspace{-2pt}
\end{table}
\subsection{Ablation Study}
\vspace{-5pt}

To analyze the contributions of the pretrained vision prior, architectural adaptations, and temporal reasoning, we conduct comprehensive ablations evaluated zero-shot on Sintel and KITTI. All variants are trained under identical zero-shot settings above to ensure fair comparisons. We also report parameter counts, latency, and peak memory to provide full transparency on computational trade-offs. 

\noindent\textbf{The Role of Priors and Scale.} 
Tab.~\ref{tab:ablation_arch} shows that scaling transformer depth $L$ from 6 to 12 layers from scratch actually degrades zero-shot performance. This indicates that without proper initialization, increasing model capacity exacerbates optimization difficulties on large displacements. However, introducing the VGGT pretrained prior effectively regularizes the training and unlocks the ability to scale up the architecture, enabling our 24-layer full model to fully leverage its massive capacity and achieve optimal performance. Meanwhile, the image encoding strategy impacts motion estimation. While a standard convolutional embedder provides reasonable local features, it lacks the aligned semantic space inherent in foundation models.

\noindent\textbf{Feature Fusion and Fine Tuning.} 
Freezing the backbone or removing the CNN feature fusion causes severe performance drops (e.g., Fl-all 10.9 to 14.7). While the prior provides a global geometric information, fine tuning and structural fusion remain essential to recover the local details.
\begin{figure}[t]
    \centering    \includegraphics[width=\linewidth]{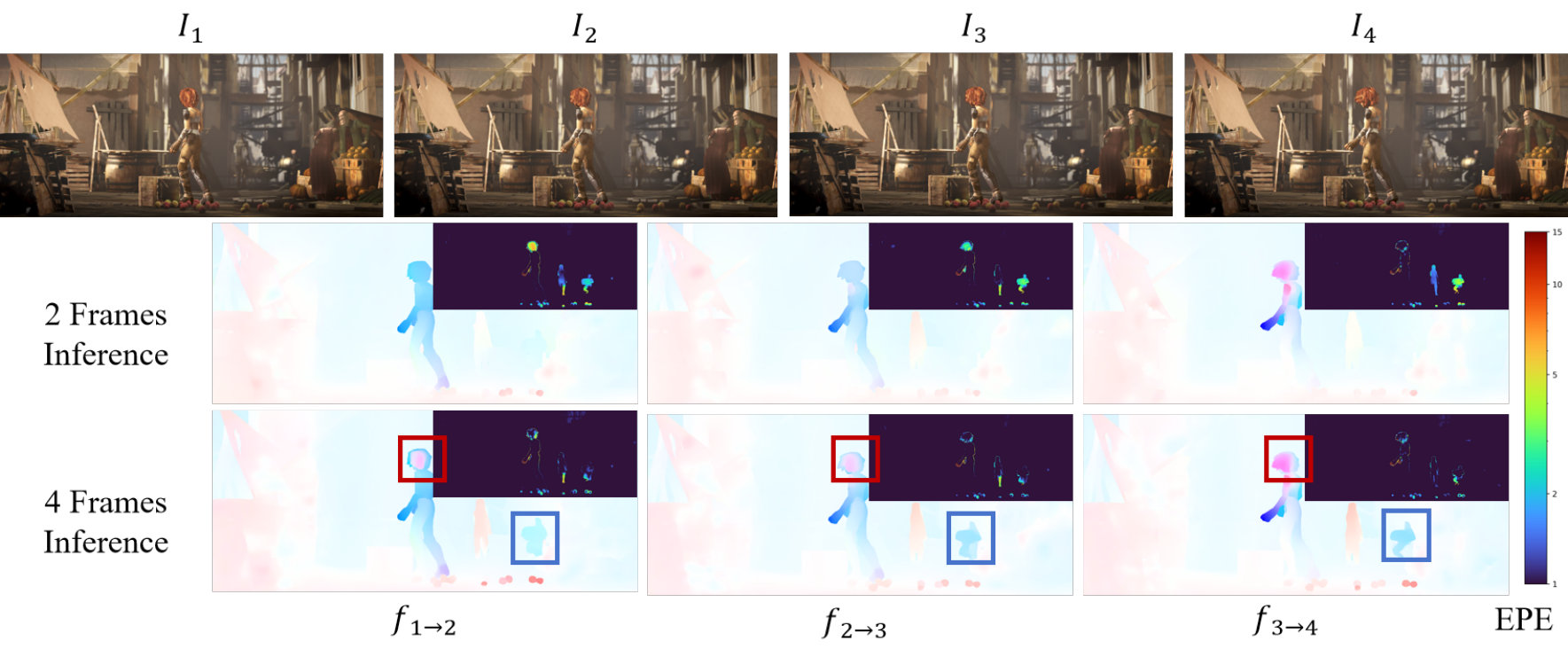}
    \caption{\textbf{Impact of multi-frame context on temporal consistency.} Top row: Consecutive input frames. Middle row: Optical flow estimated from isolated frame pairs. Bottom row: Flow estimated jointly ($T=4$). Processing isolated pairs leads to temporal inconsistencies and occlusion artifacts, particularly around the moving subject (red boxes) and background structures (blue boxes). In contrast, our expanded multi-frame context produces highly stable and accurate motion boundaries.}
    \label{fig:temp_vis}
\end{figure}
\begin{table}[t]
    \centering
    \caption{\textbf{Ablation of Temporal Attention.} Evaluated zero-shot across varying input frame counts ($T \in \{2, 4, 6\}$). Expanding the context window to $T=4$ optimally resolves complex motion blur on Sintel, while a standard 2-frame setup avoids occlusion noise on the fast, forward-driving KITTI dataset.}
    \label{tab:ablation_temporal}
    \renewcommand{\arraystretch}{1.1} 
    \vspace{-4pt}
    \setlength{\tabcolsep}{4pt} %
    \begin{tabular}{l ccc ccc ccc}
    \toprule
    \multirow{2}{*}{Method} & \multicolumn{3}{c}{Sintel Clean $\downarrow$} & \multicolumn{3}{c}{Sintel Final $\downarrow$} & \multicolumn{3}{c}{KITTI Fl-epe $\downarrow$} \\
    \cmidrule(lr){2-4} \cmidrule(lr){5-7} \cmidrule(lr){8-10}
    & 2 & 4 & 6 & 2 & 4 & 6 & 2 & 4 & 6 \\
    \midrule
    w/o temporal attn & \textbf{0.88} & 0.95 & 0.98 & 2.09 & 1.99 & 2.04 & 3.12 & 4.22 & 4.56 \\
    w/ temporal attn & 0.89 & \textbf{0.85} & \textbf{0.94} & \textbf{2.07} & \textbf{1.83} & \textbf{1.92} & \textbf{3.00} & \textbf{3.20} & \textbf{3.67} \\
    \bottomrule
    \end{tabular}
\end{table}

\noindent\textbf{Multi-Frame Consistency and Temporal Attention.}
As shown in Fig.~\ref{fig:temp_vis}, evaluating frames in isolated pairs (middle row) yields temporally inconsistent predictions and localized artifacts around occluders. By leveraging multi-frame context (bottom row), \methodname ensures highly stable flow estimations. 
Tab.~\ref{tab:ablation_temporal} quantifies this advantage. On the complex Sintel benchmark, expanding the context window to 4 frames drastically reduces the Final EPE from 2.08 to 1.83. Crucially, this improvement strictly relies on our temporal attention module. Without it, the network struggles to effectively aggregate the extended context, yielding marginal gains. 

Conversely, KITTI performance degrades as frame count increases. KITTI features rapid forward ego-motion where objects quickly exit the field of view. In these extreme scenarios, extended temporal windows introduce boundary occlusion artifacts rather than useful structural priors. Consequently, our architecture elegantly supports flexible inputs, defaulting to a 4-frame context for complex general motion while maintaining an optimal 2-frame setup for rapid ego-motion.

%% file: sec/5_conclusion.tex
\section{Conclusion}
\label{sec:conclusion}
\vspace{-8pt}

We presented \methodname, a unified architecture designed for large displacement motion estimation. Our approach demonstrates that integrating pretrained foundation vision priors with a synergistic global and local feature formulation effectively addresses the challenge of large displacements. Extensive evaluations show that \methodname achieves state-of-the-art zero-shot performance in optical flow. Crucially, this robust representation exhibits remarkable cross-task transferability. Without any architectural modifications, \methodname delivers zero-shot point tracking that rivals dedicated pipelines, and establishes new state-of-the-art results after fine-tuning on point tracking.

\noindent\textbf{Limitations and Future Work.} While \methodname demonstrates strong generalization, dense multi-frame modeling inherently increases the computational overhead for longer sequences. Future work will focus on improving sequence-level efficiency and exploring unified pre-training paradigms to jointly optimize dense optical flow and long-range tracking. Ultimately, \methodname represents a promising step toward building a robust and generalized foundation for extreme motion estimation across diverse real-world applications.

%% file: suppl.tex
\setcounter{table}{0}
\renewcommand{\thetable}{S\arabic{table}}
\setcounter{figure}{0}
\renewcommand{\thefigure}{S\arabic{figure}}
\setcounter{equation}{0}
\renewcommand{\theequation}{S\arabic{equation}}

In this supplementary material, we provide additional quantitative and qualitative results, extended ablation studies and additional implementation details that complement the main paper. 

\section{Additional Large Displacement Results}

To further evaluate the generalization capabilities of our approach, we present additional zero-shot point tracking comparisons on the TAP-Vid \cite{doersch2022tap} and RoboTAP \cite{vecerik2024robotap} benchmarks in Tab.~\ref{tab:suppl_flow_track}. Among models trained exclusively on optical flow datasets, \methodname establishes state-of-the-art average position accuracy ($\delta_{avg}$). Notably, it significantly outperforms recent strong baselines, including WAFT \cite{wang2025waft} and MemFlow-T \cite{dong2024memflow} on the zero-shot point tracking benchmarks.

\begin{figure*}[ht]
    \centering
    \includegraphics[width=\linewidth]{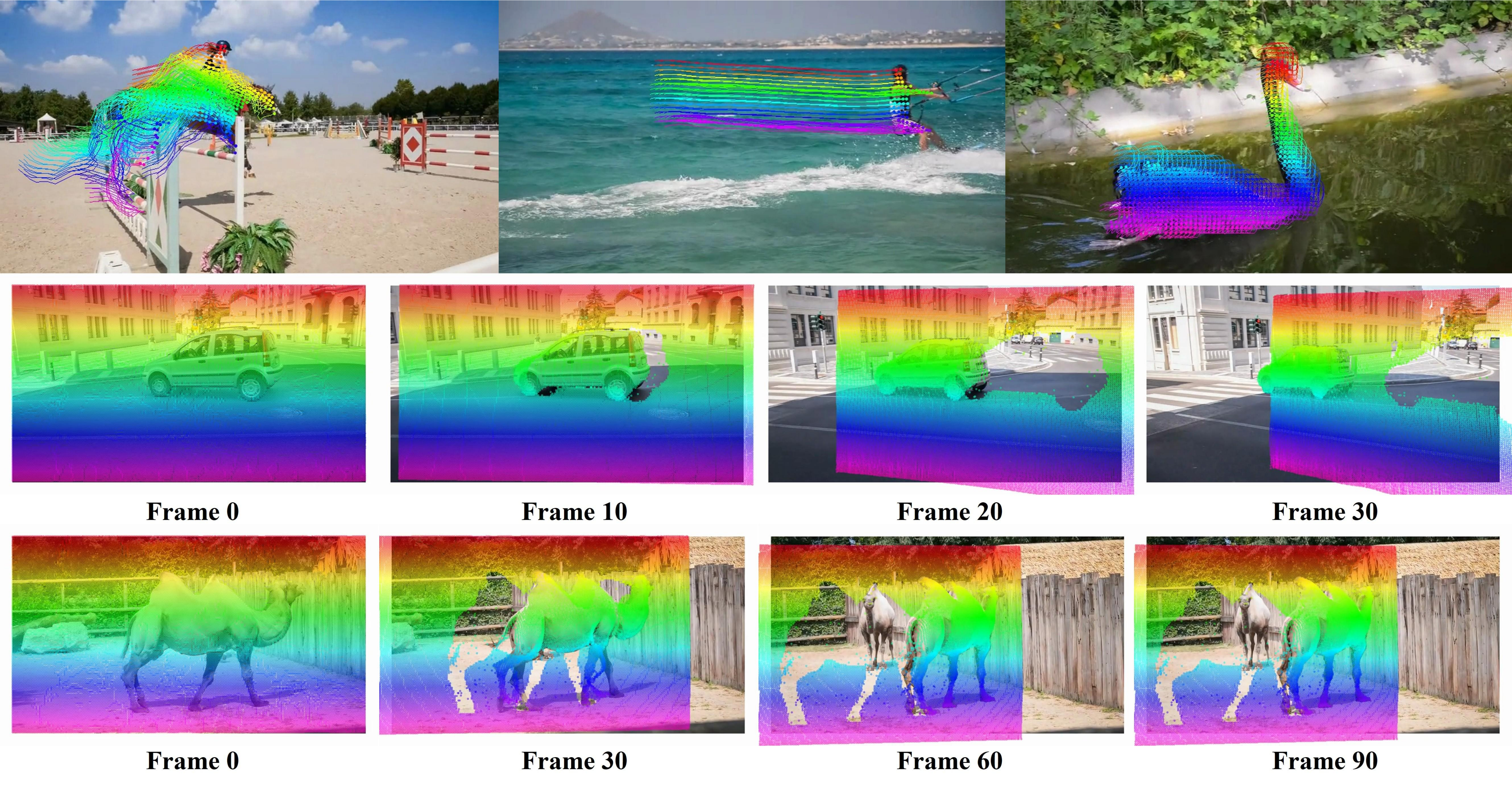}
    \caption{\textbf{Zero-shot long-range tracking visualizations.} After TSHK-stage training, our method achieves strong zero-shot tracking on the TAP-Vid \cite{doersch2022tap} dataset, enabling both dense tracking and stable long-range point trajectories.}
    \label{fig:more_track}
\end{figure*}

\begin{figure*}[ht]
    \centering
    \includegraphics[width=\linewidth]{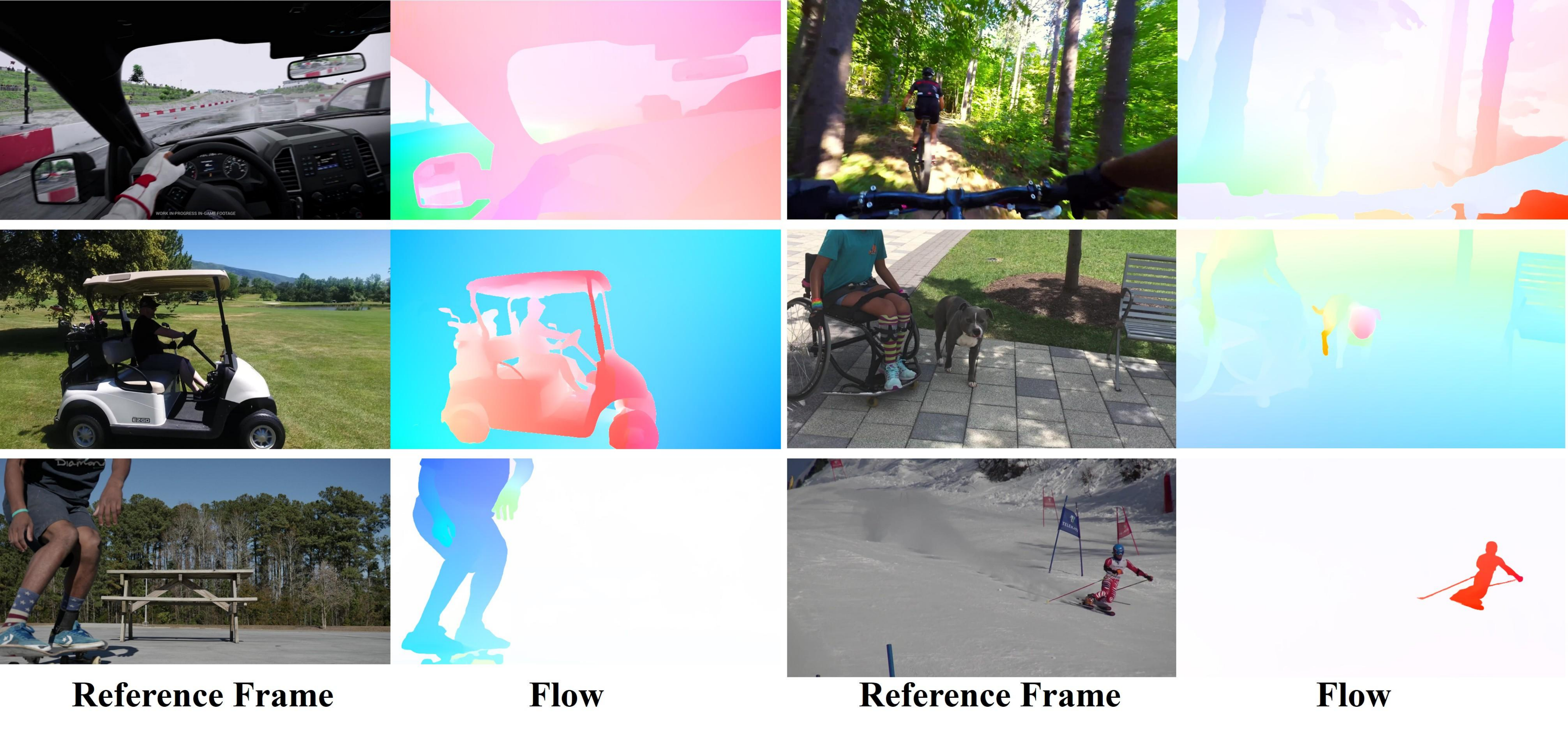}
    \caption{\textbf{Zero-shot flow estimation on DAVIS \cite{perazzi2016benchmark}.} After TSHK-stage training, our method generalizes well to diverse video scenes, demonstrating stable large displacement optical flow prediction. }
    \label{fig:davis_res}
\end{figure*}

We provide additional qualitative long-range trajectory visualizations in Fig.~\ref{fig:more_track}. Following the protocol in the main paper, we estimate the optical flow from the first frame to every subsequent frame and initialize a $1/8$-resolution point grid to track trajectories. Although trained solely on standard optical flow datasets with short temporal windows, our method generalizes robustly to long-range tracking scenarios exceeding 90 frames. Finally, Fig.~\ref{fig:davis_res} demonstrates the zero-shot applicability of our approach across diverse, complex scenes from the DAVIS dataset \cite{perazzi2016benchmark}.


\section{Additional Efficiency Analysis}
\begin{figure}[htbp]
    \centering
    \begin{minipage}[c]{0.48\linewidth}
        \centering
        \footnotesize
        \setlength{\tabcolsep}{4pt}
        \begin{tabular}{l|cccc}
        \toprule
        Method & P & F & L & M \\
        \midrule
        Flowformer++ \cite{shi2023flowformer++} & 18.2 & 2 & 211 & 2.1 \\
        SAMFlow \cite{zhou2024samflow} & 659 & 2 & 438 & 5.4 \\
        FlowDiffuser \cite{luo2024flowdiffuser} & 14.5 & 2 & 465 & 5.1 \\
        VideoFlow \cite{shi2023videoflow} & 13.6 & 5 & 383 & 4.2 \\
        MemFlow-T \cite{dong2024memflow} & 12.7 & 3 & 171 & 2.0 \\
        UFM \cite{zhang2025ufm} & 478 & 2 & 306 & 4.1 \\
        WAFT \cite{wang2025waft} & 56.5 & 2 & 106 & 1.2 \\
        \midrule
        MegaFlow-S & 256 & 2 & 148 & 2.9 \\
        MegaFlow-M & 483 & 2 & 210 & 4.3 \\
        MegaFlow & 936 & 4 & 273 & 7.5 \\
        MegaFlow & 936 & 2 & 328 & 6.1 \\
        \bottomrule
        \end{tabular}
        \makeatletter\def\@captype{table}\makeatother
        \caption{\textbf{Efficiency comparison.} P: Params (M); F: Frames; L: latency per output flow (ms); M: peak memory (GB). Measured on an RTX 3090 at $540 \times 960$.}
        \label{tab:efficiency}
    \end{minipage}\hfill
    \begin{minipage}[c]{0.5\linewidth}
        \centering
        \includegraphics[width=\linewidth]{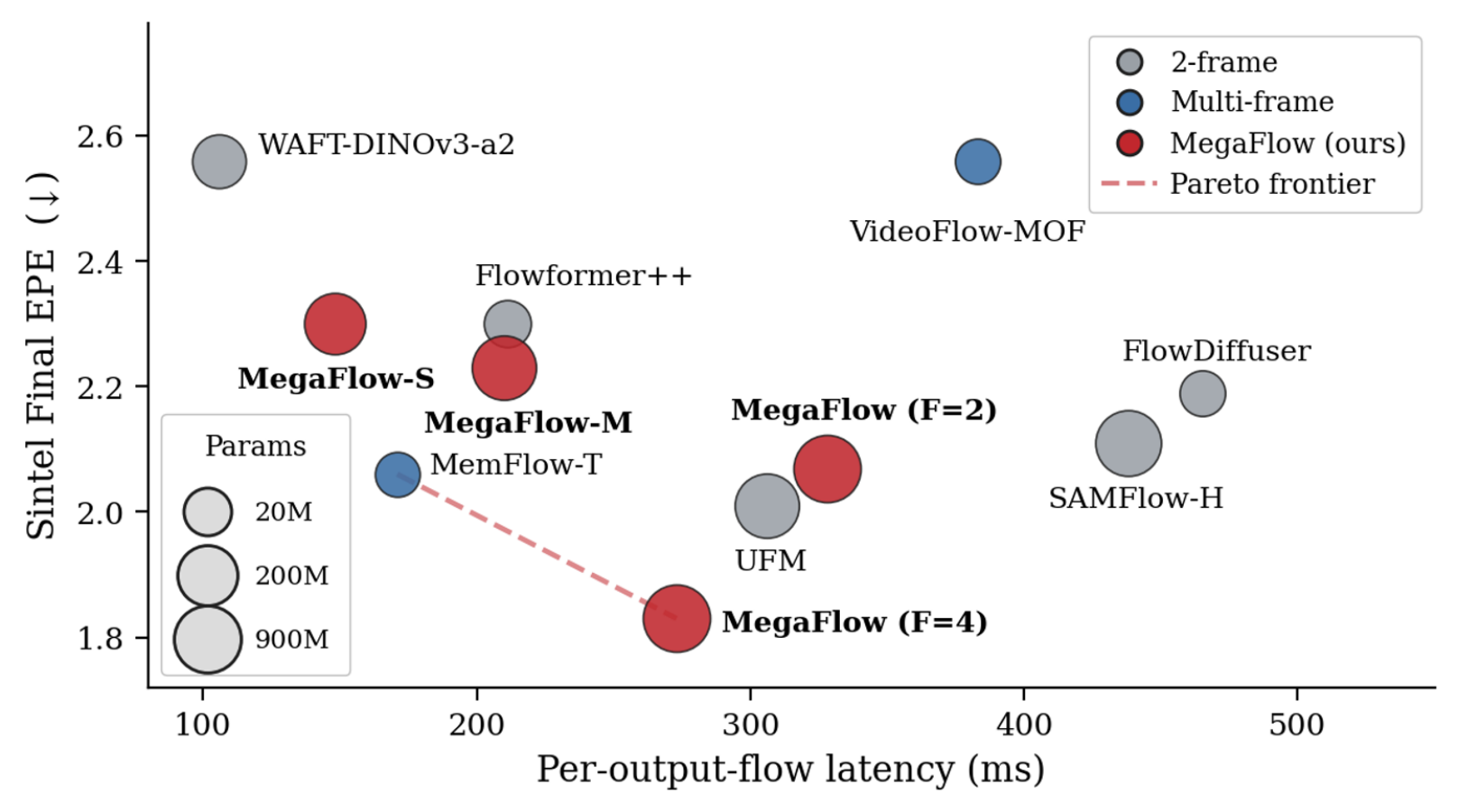}
        \caption{\textbf{Pareto frontier.} \methodname{} establishes a new state-of-the-art trade-off for zero-shot accuracy versus latency.}
        \label{fig:efficiency}
    \end{minipage}
\end{figure}

Tab.~\ref{tab:efficiency} and Fig.~\ref{fig:efficiency}(a) provide detailed efficiency comparisons evaluated on a single RTX 3090 GPU using matched bf16 precision. Notably, the per-pair inference latency decreases as the frame count $T$ increases. This occurs because our alternating backbone successfully amortizes the computational cost of global matching across the extended temporal window. As a result, \methodname{} (936M parameters, $T=4$) operates 1.6 times faster than SAMFlow (659M parameters), while simultaneously maintaining superior accuracy across all evaluated metrics. This inference efficiency is driven by our architectural design, which relies on coarse global matching and lightweight local refinement, reducing the required number of recurrent iterations despite the larger overall parameter count.

Fig~\ref{fig:efficiency} demonstrates that \methodname{} establishes a new Pareto frontier at the high-accuracy spectrum on the zero-shot Sintel Final benchmark. To demonstrate scalability, we also evaluate \methodname-S and \methodname-M, which are initialized from the first 6 and 12 VGGT blocks respectively in Tab.~\ref{tab:ablation_arch} while keeping all other components identical. These variants reach competitive performance using approximately 4 GH200-days of compute, and both consistently outperform WAFT and prior baselines on the Sintel Final zero-shot benchmark. We acknowledge the training computational overhead as a limitation and intend to explore more parameter-efficient training strategies in future work.

\begin{table}[t]
    \centering
    \caption{\textbf{Zero-shot point tracking comparison} on the TAP-Vid \cite{doersch2022tap} and RoboTAP \cite{vecerik2024robotap} benchmarks. We evaluate the average position accuracy ($\delta_{avg}$) using an input resolution of $384 \times 512$. Among models trained exclusively on optical flow datasets, \methodname establishes new state-of-the-art results.}
    \label{tab:suppl_flow_track}
    \vspace{-4pt}
    
    \begin{tabular*}{\textwidth}{@{\extracolsep{\fill}} l c c c c c @{}}
    \toprule
    Method & DAVIS $\uparrow$ & Kinetics $\uparrow$ & RGB-Stacking $\uparrow$ & RoboTAP $\uparrow$ & Mean $\uparrow$ \\
    \midrule
    RAFT~\cite{teed2020raft} &  48.5 & 64.3 & 82.8 & 72.2 & 67.0 \\
    SEA-RAFT~\cite{wang2024sea} &  48.7 & 64.3 & 85.7 & 67.6 & 66.6 \\
    AccFlow~\cite{wu2023accflow} &  23.5 & 38.8 & 63.2 & 57.9 & 45.9 \\
    MemFlow-T~\cite{dong2024memflow} & 61.7 & 64.9 & 85.2 & 77.0 & 72.2 \\
    WAFT-DINOv3-a2~\cite{wang2025waft} & 53.9 & 61.0 & 84.0 & 73.0 & 68.0 \\
    \methodname &  \textbf{65.6} & \textbf{65.5} & \textbf{89.6} & \textbf{77.5} & \textbf{74.6} \\
    \bottomrule
    \end{tabular*}
    \vspace{-2mm}
\end{table}
\begin{table}[t]
\centering
\caption{\textbf{Ablation Studies.} All models are evaluated in a zero-shot setting on Sintel and KITTI training datasets using 4-frame inputs. (a) Impact of progressive pretraining stages. (b) Trade-off between accuracy and inference latency across different refinement iterations ($K$). Latency is measured in ms per flow on an RTX 4090.}
\begin{tabular}{l cc cc c}
\toprule
\multirow{2}{*}{Setting} & \multicolumn{2}{c}{Sintel (train)} & \multicolumn{2}{c}{KITTI (train)} & Latency \\
\cmidrule(l{0.5ex}r{0.5ex}){2-3} \cmidrule(l{0.5ex}r{0.5ex}){4-5}
& Clean$\downarrow$ & Final$\downarrow$ & Fl-epe$\downarrow$ & Fl-all$\downarrow$ & (ms)$\downarrow$ \\ 
\midrule
\multicolumn{6}{l}{\textit{(a) Training Stages}} \\
w/o TartanAir & 0.94 & 1.90 & 3.78 & 12.10 & 169.2 \\
w/o Things-2f & 0.88 & 1.88 & \textbf{3.39} & 12.20 & 169.2 \\
\methodname (Full) & \textbf{0.85} & \textbf{1.83} & 3.44 & \textbf{11.10} & 169.2 \\
\midrule
\multicolumn{6}{l}{\textit{(b) Refinement Iterations ($K$)}} \\
$K=1$ & 1.07 & 2.02 & 5.92 & 19.03 & 107.4 \\
$K=2$ & 0.95 & 1.92 & 5.12 & 15.56 & 115.7 \\
$K=4$ & 0.88 & 1.86 & 4.11 & 12.26 & 134.4 \\
$K=6$ & 0.87 & 1.85 & 3.65 & 11.38 & 149.7 \\
$K=8$ (Default) & \textbf{0.85} & \textbf{1.83} & 3.44 & 11.14 & 169.2 \\
$K=10$ & 0.88 & 1.85 & 3.29 & 10.94 & 183.8 \\
$K=12$ & 0.88 & 1.86 & \textbf{3.18} & \textbf{10.81} & 205.0 \\
\bottomrule
\end{tabular}
\vspace{-5pt}
\label{tab:supp_ablations}
\end{table}

\section{Additional Ablations}

This section provides further ablation studies to supplement the main manuscript.

\noindent \textbf{Effect of Different Training Stages.}
We further investigate the impact of the TartanAir~\cite{tartanair2020iros} and 2-frame FlyingThings (Things-2f) \cite{flyingthings} pretraining stages on our zero-shot performance. As detailed in Tab.~\ref{tab:train}, our pipeline progressively scales the temporal context. Tab.~\ref{tab:supp_ablations} (a) presents the ablation results. 
Removing the TartanAir stage (\textit{w/o TartanAir}) leads to a noticeable performance drop across all benchmarks. Specifically, Sintel Clean error increases from 0.85 to 0.94, and the KITTI Fl-all outlier metric degrades from 11.1 to 12.1. This demonstrates that the diverse camera trajectories and rigid structures in TartanAir provide a robust geometric prior that broadly benefits both synthetic and real-world generalization.
Similarly, bypassing the 2-frame FlyingThings warmup (\textit{w/o Things-2f}) hurts overall robustness. While KITTI Fl-epe sees a minor artifactual drop without it, the crucial KITTI Fl-all outlier metric worsens significantly from 11.1 to 12.2. Furthermore, both Sintel Clean and Final errors increase without this stage. This confirms that a sequential curriculum, smoothly transitioning from 2-frame to multi-frame inputs, stabilizes the optimization process and effectively suppresses large outlier predictions. Ultimately, the full \methodname achieves the best balance and sets a highly competitive zero-shot baseline.

\noindent \textbf{Effect of Refinement Iterations.} We investigate the impact of the number of refinement iterations ($K$) on both accuracy and inference latency, as shown in Tab.~\ref{tab:supp_ablations} (b). Increasing $K$ from 1 to 8 consistently reduces prediction errors across both benchmarks. At $K=8$, the model achieves its optimal performance on Sintel while maintaining a reasonable latency of 169.2 ms per flow. Pushing the iterations further ($K > 8$) yields only marginal improvements on KITTI while slightly degrading Sintel accuracy and significantly increasing computational cost. Therefore, we select $K=8$ as our default configuration to strike the best balance between peak accuracy and inference efficiency.

\section{Implementation Details}
\noindent \textbf{Architecture.}
As described in the main paper, \methodname adopts a 24-block transformer backbone, where each block contains a frame-wise self-attention layer and a global self-attention layer following the design of VGGT~\cite{wang2025vggt}. The architecture follows the ViT-L configuration used in DINOv2~\cite{oquab2023dinov2}, with a feature dimension of 1024 and 16 attention heads. We use the official PyTorch implementation with FlashAttention-3~\cite{shah2024flashattention} enabled for efficient training. Image tokenization is performed by the DINOv2 encoder with positional embeddings added to the tokens. Similar to ~\cite{depth_anything_v2, wang2025vggt, keetha2025mapanything}, tokens from the 4th, 11th, 17th, and 23rd layers are fed into a DPT-style decoder~\cite{ranftl2021vision} to generate a $1/7$ resolution feature map. We modify the final DPT resize layer to match the preceding layer’s configuration, allowing two consecutive upsampling steps and producing a maximum resolution of $1/7$. 

To fuse high-resolution CNN features, we incorporate 1/2 and 1/4 feature maps from the CNN encoder. Both feature maps are first downsampled to $1/8$ resolution using pixel unshuffle \cite{shi2016real} and then bilinearly resized to $1/7$ resolution. These aligned CNN features are concatenated with the transformer features and passed through lightweight $1 \times 1$ convolutional fusion layers, producing a unified $1/7$ feature representation for each frame $F_i \in \mathbb{R}^{128 \times H/7 \times W/7}$. 

In the local refinement stage, we iteratively update the optical flow estimate $f_i$ for each frame. Given the local correlation volume ${C}_i^{\text{local}} \in \mathbb{R}^{(2r+1)^2 \times H/4 \times W/4}$ and the fused feature map $F_i$, we first extract motion features ${M}_i$ using a convolutional motion encoder, adopting the design from~\cite{teed2020raft, wang2024sea}. These features are subsequently fed into hybrid spatiotemporal refinement blocks, along with the current hidden state ${h}_i$ and context features $F_i$. The hidden state $h_i$ first undergoes a spatial update via the RNN cell \cite{wang2024sea}. Subsequently, we perform attention along the temporal dimension to capture motion consistency across frames. Finally, the residual flow $\Delta f_i$ is regressed from the updated state $h_i''$ via a two-layer convolutional head: 
\begin{align} 
{M}_i &= \text{MotionEncoder}({C}_i^{\text{local}}, f_i) \\ 
{h}_i' &= \text{RNN}({h}_i, {M}_i, F_i) \\ 
{h}_i'' &= \text{TempAttn}(\text{Reshape}({h}_i')) \\ 
\Delta f_i &= \text{FlowHead}({h}_i'') 
\end{align}

Specifically, we set the radius of the local window as $r=4$. We employ $4$ update iterations during the training phase, while allowing up to $8$ iterations during inference.

\begin{table}[htbp]
    \caption{\textbf{Training details for each stage.} 
    Dataset abbreviations: Chairs = FlyingChairs \cite{dosovitskiy2015flownet}, TartanAir = TartanAirV1\cite{tartanair2020iros}, T = FlyingThings~\cite{flyingthings}, 
    S = Sintel~\cite{butler2012naturalistic}, 
    K = KITTI-2015~\cite{geiger2012we}, 
    H = HD1K~\cite{kondermann2016hci} and Kubric \cite{greff2022kubric}. 
    Frames denotes the number of input frames used during training. 
    \texttt{randn}(m, n) indicates that the number of input frames is randomly sampled between $m$ and $n$.}
    \centering
    \begin{tabular}{lccccccc}
    \toprule
       Stage  & Weights & Datasets & Crop Size & LR & Batch Size & Steps & Frames \\
    \midrule
        Chairs & - & Chairs & [384, 512] & 5e-4 & 128 & 20k & 2 \\
        TartanAir & Chairs & TartanAir & [480, 640] & 2e-4 & 128 & 30k & 2 \\
        Things-2f & TartanAir & T & [432, 960] & 2e-4 & 128 & 15k & 2 \\
        Things & Things-2f & T & [432, 960] & 2e-4 & 128 & 15k & \texttt{randn}(2,6) \\
        TSHK & Things & T+S+H+K & [432, 960] & 2e-4 & 128 & 30k & \texttt{randn}(2,6) \\
        Kubric & TSHK & Kubric & [384, 512] & 2e-4 & 128 & 20K & 16 \\
    \bottomrule
    \end{tabular}
    \vspace{-5pt}
    \label{tab:train}
\end{table}

\noindent \textbf{Training.}
We employ the AdamW~\cite{loshchilov2017decoupled} optimizer with different learning rates for different components of the model. The DINOv2 tokenizer is frozen, and the learning rate for the alternating transformer blocks is set to be $100\times$ smaller than that for the remaining parameters. We use a cosine annealing learning rate scheduler~\cite{smith2019super} during training. Detailed training configurations for each stage are provided in Tab.~\ref{tab:train}. Since KITTI only provides ground-truth flow for a single frame, we follow its multi-view extension \cite{geiger2012we} to supply multi-frame inputs while supervising flow on only one frame.

\begin{figure*}[h]
    \centering
    \includegraphics[width=\linewidth]{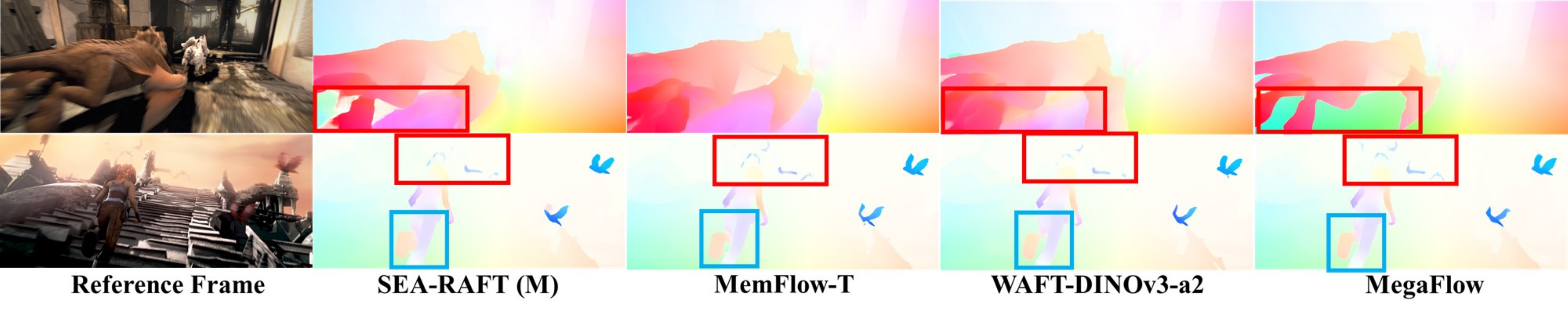}
  
    \caption{\textbf{Qualitative comparison of optical flow on Sintel.} Visualization of SEA-RAFT \cite{wang2024sea}, MemFlow \cite{dong2024memflow}, WAFT-DINOv3-a2 \cite{wang2025waft}, and our method on the Sintel Final test set.}

    \label{fig:sintel-comp}
\end{figure*}

\section{Additional Benchmark Results}
We provide extended benchmark comparisons to further evaluate the generalization capability of our approach. Table~\ref{tab:sintel-res} reports per-sequence results on the Sintel benchmark. Consistent with prior observations \cite{wang2025waft, saxena2023surprising}, the \texttt{Ambush 1} sequence remains an outlier for our method under extreme blur.

\begin{figure}[htbp]
    \centering
    \includegraphics[width=\linewidth]{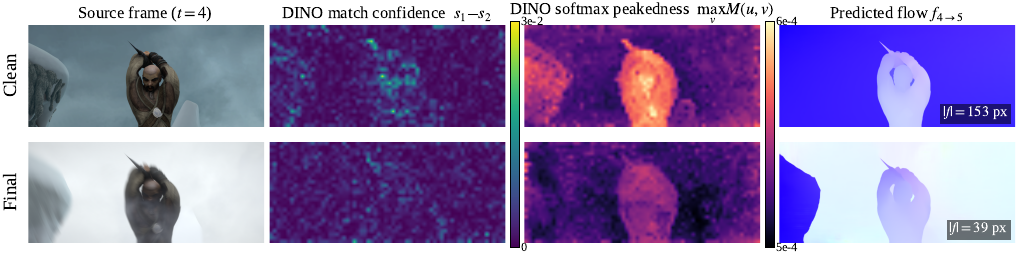}
    \caption{\textbf{Failure analysis on the \texttt{Ambush 1} sequence.} Extreme blur in the Final pass degrades features, which flattens the global softmax distribution and collapses the initial displacement estimate.}
    \label{fig:ambush}
\end{figure}

To analyze this specific failure mode, we compare our Clean versus Final estimates (Figure~\ref{fig:ambush}) since ground-truth flow is unavailable. On the Sintel Clean pass, the DINO features remain discriminative; our global matching yields a peaked distribution and successfully recovers the large 153px displacement. However, on the Final pass, severe blur degrades the DINO features. This degradation flattens the softmax distribution, biases the expectation toward the candidate mean, and collapses the initialization to just 39px. Subsequent local refinement cannot bridge such a massive gap. This failure is specific to our global softmax step operating on severely degraded features. While iterative local refinement architectures like VideoFlow \cite{shi2023videoflow} are less affected by this specific issue, they lack our structural advantage for large displacements. When excluding \texttt{Ambush 1} (Table~S4), \methodname{} remains highly competitive with VideoFlow \cite{shi2023videoflow} and WAFT \cite{wang2025waft}, and outperforms other baselines on Sintel Final.

Figure~\ref{fig:sintel-comp} provides qualitative comparisons on the Sintel benchmark. Compared to recent approaches, \methodname{} demonstrates superior robustness in handling complex occlusions and severe motion blur in most scenarios, consistently producing sharper and more accurate motion boundaries.

\begin{table*}[t]
    \centering
    \caption{Detailed Endpoint-error (EPE) on Sintel~\cite{butler2012naturalistic} sequences for the Final pass. We report the average across all sequences, as well as the average excluding the extreme outlier sequence (Ambush 1).}
    \resizebox{1.0\linewidth}{!}{
    \begin{tabular}{lccccc}
    \toprule
        Sequence & \methodname & WAFT-DINOv3-a2 & SAMFlow & MemFlow-T & VideoFlow-BOF \\
    \midrule
        Perturbed Market 3 & 0.873 & 0.788 & 0.932 & 0.936 & 0.834 \\
        Perturbed Shaman 1 & 0.294 & 0.180 & 0.241 & 0.228 & 0.228 \\
        Ambush 1 & 24.069 & 16.161 & 10.586 & 7.362 & 6.461 \\
        Ambush 3 & 3.003 & 3.012 & 3.411 & 2.678 & 2.710 \\
        Bamboo 3 & 0.469 & 0.467 & 0.522 & 0.524 & 0.477 \\
        Cave 3 & 1.833 & 1.978 & 2.475 & 2.176 & 1.952 \\
        Market 1 & 0.850 & 0.836 & 1.060 & 0.872 & 0.781 \\
        Market 4 & 6.061 & 6.126 & 6.636 & 7.171 & 6.189 \\
        Mountain 2 & 0.422 & 0.398 & 0.500 & 0.671 & 0.451 \\
        Temple 1 & 0.477 & 0.520 & 0.853 & 0.527 & 0.539 \\
        Tiger & 0.480 & 0.452 & 0.573 & 0.494 & 0.490 \\
        Wall & 2.357 & 1.330 & 2.105 & 1.640 & 1.517 \\
    \midrule
        Average & 3.432 & 2.687 & 2.491 & 2.107 & \textbf{1.886} \\
        Average w/o Ambush 1 & 1.556 & \textbf{1.462} & 1.755 & 1.629 & 1.470 \\
    \bottomrule
    \end{tabular}
    }
    \label{tab:sintel-res}
\end{table*}

%% file: main.bib
@String(CVPR  = {IEEE Conf. Comput. Vis. Pattern Recog.})

@String(ICCV  = {Int. Conf. Comput. Vis.})

@String(ECCV  = {Eur. Conf. Comput. Vis.})

@String(AAAI  = {AAAI})

@String(IJCAI = {IJCAI})

@String(CVPR  = {CVPR})

@String(ICCV  = {ICCV})

@String(ECCV  = {ECCV})

@article{wang2024sea,
  title={SEA-RAFT: Simple, Efficient, Accurate RAFT for Optical Flow},
  author={Wang, Yihan and Lipson, Lahav and Deng, Jia},
  journal={arXiv preprint arXiv:2405.14793},
  year={2024}
}

@inproceedings{teed2020raft,
  title={Raft: Recurrent all-pairs field transforms for optical flow},
  author={Teed, Zachary and Deng, Jia},
  booktitle={European conference on computer vision},
  pages={402--419},
  year={2020},
  organization={Springer}
}

@inproceedings{harley2025alltracker,
author    = {Adam W. Harley and Yang You and Xinglong Sun and Yang Zheng and Nikhil Raghuraman and Yunqi Gu and Sheldon Liang and Wen-Hsuan Chu and Achal Dave and Pavel Tokmakov and Suya You and Rares Ambrus and Katerina Fragkiadaki and Leonidas J. Guibas},
title     = {All{T}racker: {E}fficient Dense Point Tracking at High Resolution},
booktitle = {ICCV},
year      = {2025}
}

@article{bargatin2025memfof,
  title={MEMFOF: High-Resolution Training for Memory-Efficient Multi-Frame Optical Flow Estimation},
  author={Bargatin, Vladislav and Chistov, Egor and Yakovenko, Alexander and Vatolin, Dmitriy},
  journal={arXiv preprint arXiv:2506.23151},
  year={2025}
}

@inproceedings{wu2023accflow,
  title={Accflow: Backward accumulation for long-range optical flow},
  author={Wu, Guangyang and Liu, Xiaohong and Luo, Kunming and Liu, Xi and Zheng, Qingqing and Liu, Shuaicheng and Jiang, Xinyang and Zhai, Guangtao and Wang, Wenyi},
  booktitle={Proceedings of the IEEE/CVF International Conference on Computer Vision},
  pages={12119--12128},
  year={2023}
}

@inproceedings{zheng2023point,
    author = {Yang Zheng and Adam W. Harley and Bokui Shen and Gordon Wetzstein and Leonidas J. Guibas},
    title = {PointOdyssey: A Large-Scale Synthetic Dataset for Long-Term Point Tracking},
    booktitle = {ICCV},
    year = {2023}
}

@article{cho2024local,
  title={Local All-Pair Correspondence for Point Tracking},
  author={Cho, Seokju and Huang, Jiahui and Nam, Jisu and An, Honggyu and Kim, Seungryong and Lee, Joon-Young},
  journal={arXiv preprint arXiv:2407.15420},
  year={2024}
}

@inproceedings{karaev2024cotracker,
  title={Cotracker: It is better to track together},
  author={Karaev, Nikita and Rocco, Ignacio and Graham, Benjamin and Neverova, Natalia and Vedaldi, Andrea and Rupprecht, Christian},
  booktitle={European conference on computer vision},
  pages={18--35},
  year={2024},
  organization={Springer}
}

@InProceedings{karaev2024cotracker3,
  title={Cotracker3: Simpler and better point tracking by pseudo-labelling real videos},
  author={Karaev, Nikita and Makarov, Yuri and Wang, Jianyuan and Neverova, Natalia and Vedaldi, Andrea and Rupprecht, Christian},
  booktitle={Proceedings of the IEEE/CVF International Conference on Computer Vision},
  pages={6013--6022},
  year={2025}
}

@inproceedings{doersch2024bootstap,
  title={Bootstap: Bootstrapped training for tracking-any-point},
  author={Doersch, Carl and Luc, Pauline and Yang, Yi and Gokay, Dilara and Koppula, Skanda and Gupta, Ankush and Heyward, Joseph and Rocco, Ignacio and Goroshin, Ross and Carreira, Joao and others},
  booktitle={Proceedings of the Asian Conference on Computer Vision},
  pages={3257--3274},
  year={2024}
}

@article{ngo2024delta,
  author    = {Ngo, Tuan Duc and Zhuang, Peiye and Gan, Chuang and Kalogerakis, Evangelos and Tulyakov, Sergey and Lee, Hsin-Ying and Wang, Chaoyang},
  title     = {DELTA: Dense Efficient Long-range 3D Tracking for Any video},
  journal   = {arXiv preprint arXiv:2410.24211},
  year      = {2024}
}

@inproceedings{lucas1981iterative,
  title={An iterative image registration technique with an application to stereo vision},
  author={Lucas, Bruce D and Kanade, Takeo},
  booktitle={IJCAI'81: 7th international joint conference on Artificial intelligence},
  volume={2},
  pages={674--679},
  year={1981}
}

@article{horn1981determining,
  title={Determining optical flow},
  author={Horn, Berthold KP and Schunck, Brian G},
  journal={Artificial intelligence},
  volume={17},
  number={1-3},
  pages={185--203},
  year={1981},
  publisher={Elsevier}
}

@inproceedings{dosovitskiy2015flownet,
  title={Flownet: Learning optical flow with convolutional networks},
  author={Dosovitskiy, Alexey and Fischer, Philipp and Ilg, Eddy and Hausser, Philip and Hazirbas, Caner and Golkov, Vladimir and Van Der Smagt, Patrick and Cremers, Daniel and Brox, Thomas},
  booktitle=ICCV,
  pages={2758--2766},
  year={2015}
}

@inproceedings{ranjan2017optical,
  title={Optical flow estimation using a spatial pyramid network},
  author={Ranjan, Anurag and Black, Michael J},
  booktitle=CVPR,
  pages={4161--4170},
  year={2017}
}

@article{wang2025waft,
  title={WAFT: Warping-Alone Field Transforms for Optical Flow},
  author={Wang, Yihan and Deng, Jia},
  journal={arXiv preprint arXiv:2506.21526},
  year={2025}
}

@article{xu2023unifying,
  title={Unifying flow, stereo and depth estimation},
  author={Xu, Haofei and Zhang, Jing and Cai, Jianfei and Rezatofighi, Hamid and Yu, Fisher and Tao, Dacheng and Geiger, Andreas},
  journal={IEEE Transactions on Pattern Analysis and Machine Intelligence},
  year={2023},
  publisher={IEEE}
}

@inproceedings{butler2012naturalistic,
  title={A naturalistic open source movie for optical flow evaluation},
  author={Butler, Daniel J and Wulff, Jonas and Stanley, Garrett B and Black, Michael J},
  booktitle=ECCV,
  pages={611--625},
  year={2012},
  organization={Springer}
}

@inproceedings{mehl2023spring,
  title={Spring: A high-resolution high-detail dataset and benchmark for scene flow, optical flow and stereo},
  author={Mehl, Lukas and Schmalfuss, Jenny and Jahedi, Azin and Nalivayko, Yaroslava and Bruhn, Andr{\'e}s},
  booktitle=CVPR,
  pages={4981--4991},
  year={2023}
}

@inproceedings{sun2018pwc,
  title={Pwc-net: Cnns for optical flow using pyramid, warping, and cost volume},
  author={Sun, Deqing and Yang, Xiaodong and Liu, Ming-Yu and Kautz, Jan},
  booktitle=CVPR,
  pages={8934--8943},
  year={2018}
}

@inproceedings{dong2024memflow,
  title={MemFlow: Optical Flow Estimation and Prediction with Memory},
  author={Dong, Qiaole and Fu, Yanwei},
  booktitle=CVPR,
  pages={19068--19078},
  year={2024}
}

@inproceedings{shi2023videoflow,
  title={Videoflow: Exploiting temporal cues for multi-frame optical flow estimation},
  author={Shi, Xiaoyu and Huang, Zhaoyang and Bian, Weikang and Li, Dasong and Zhang, Manyuan and Cheung, Ka Chun and See, Simon and Qin, Hongwei and Dai, Jifeng and Li, Hongsheng},
  booktitle=ICCV,
  pages={12469--12480},
  year={2023}
}

@article{xu2023memory,
  title={Memory-efficient optical flow via radius-distribution orthogonal cost volume},
  author={Xu, Gangwei and Chen, Shujun and Jia, Hao and Feng, Miaojie and Yang, Xin},
  journal={arXiv preprint arXiv:2312.03790},
  year={2023}
}

@inproceedings{geiger2012we,
  title={Are we ready for autonomous driving? the kitti vision benchmark suite},
  author={Geiger, Andreas and Lenz, Philip and Urtasun, Raquel},
  booktitle=CVPR,
  pages={3354--3361},
  year={2012},
  organization={IEEE}
}

@inproceedings{morimitsu2025dpflow,
  title={Dpflow: Adaptive optical flow estimation with a dual-pyramid framework},
  author={Morimitsu, Henrique and Zhu, Xiaobin and Cesar, Roberto M and Ji, Xiangyang and Yin, Xu-Cheng},
  booktitle={Proceedings of the Computer Vision and Pattern Recognition Conference},
  pages={17810--17820},
  year={2025}
}

@inproceedings{kondermann2016hci,
  title={The hci benchmark suite: Stereo and flow ground truth with uncertainties for urban autonomous driving},
  author={Kondermann, Daniel and Nair, Rahul and Honauer, Katrin and Krispin, Karsten and Andrulis, Jonas and Brock, Alexander and Gussefeld, Burkhard and Rahimimoghaddam, Mohsen and Hofmann, Sabine and Brenner, Claus and others},
  booktitle={Proceedings of the IEEE Conference on Computer Vision and Pattern Recognition Workshops},
  pages={19--28},
  year={2016}
}

@inproceedings{richter2017playing,
  title={Playing for benchmarks},
  author={Richter, Stephan R and Hayder, Zeeshan and Koltun, Vladlen},
  booktitle=ICCV,
  pages={2213--2222},
  year={2017}
}

@article{sun2025streamflow,
  title={Streamflow: streamlined multi-frame optical flow estimation for video sequences},
  author={Sun, Shangkun and Liu, Jiaming and Li, Huaxia and Liu, Guoqing and Li, Thomas and Gao, Wei},
  journal={Advances in Neural Information Processing Systems},
  volume={37},
  pages={9205--9228},
  year={2025}
}

@article{lai2026cowtracker,
  title={CoWTracker: Tracking by Warping instead of Correlation},
  author={Lai, Zihang and Insafutdinov, Eldar and Sucar, Edgar and Vedaldi, Andrea},
  journal={arXiv preprint arXiv:2602.04877},
  year={2026}
}

@inproceedings{weinzaepfel2023croco,
  title={Croco v2: Improved cross-view completion pre-training for stereo matching and optical flow},
  author={Weinzaepfel, Philippe and Lucas, Thomas and Leroy, Vincent and Cabon, Yohann and Arora, Vaibhav and Br{\'e}gier, Romain and Csurka, Gabriela and Antsfeld, Leonid and Chidlovskii, Boris and Revaud, J{\'e}r{\^o}me},
  booktitle=ICCV,
  pages={17969--17980},
  year={2023}
}

@inproceedings{jiang2021learning_GMA,
  title={Learning to estimate hidden motions with global motion aggregation},
  author={Jiang, Shihao and Campbell, Dylan and Lu, Yao and Li, Hongdong and Hartley, Richard},
  booktitle=ICCV,
  pages={9772--9781},
  year={2021}
}

@inproceedings{xu2022gmflow,
  title={Gmflow: Learning optical flow via global matching},
  author={Xu, Haofei and Zhang, Jing and Cai, Jianfei and Rezatofighi, Hamid and Tao, Dacheng},
  booktitle=CVPR,
  pages={8121--8130},
  year={2022}
}

@inproceedings{shi2023flowformer++,
  title={Flowformer++: Masked cost volume autoencoding for pretraining optical flow estimation},
  author={Shi, Xiaoyu and Huang, Zhaoyang and Li, Dasong and Zhang, Manyuan and Cheung, Ka Chun and See, Simon and Qin, Hongwei and Dai, Jifeng and Li, Hongsheng},
  booktitle=CVPR,
  pages={1599--1610},
  year={2023}
}

@inproceedings{huang2022flowformer,
  title={Flowformer: A transformer architecture for optical flow},
  author={Huang, Zhaoyang and Shi, Xiaoyu and Zhang, Chao and Wang, Qiang and Cheung, Ka Chun and Qin, Hongwei and Dai, Jifeng and Li, Hongsheng},
  booktitle=ECCV,
  pages={668--685},
  year={2022},
  organization={Springer}
}

@article{saxena2023surprising,
  title={The surprising effectiveness of diffusion models for optical flow and monocular depth estimation},
  author={Saxena, Saurabh and Herrmann, Charles and Hur, Junhwa and Kar, Abhishek and Norouzi, Mohammad and Sun, Deqing and Fleet, David J},
  journal={Advances in Neural Information Processing Systems},
  volume={36},
  pages={39443--39469},
  year={2023}
}

@inproceedings{morimitsu2024recurrent,
  title={Recurrent partial kernel network for efficient optical flow estimation},
  author={Morimitsu, Henrique and Zhu, Xiaobin and Ji, Xiangyang and Yin, Xu-Cheng},
  booktitle=AAAI,
  volume={38},
  pages={4278--4286},
  year={2024}
}

@inproceedings{zheng2022dip,
  title={Dip: Deep inverse patchmatch for high-resolution optical flow},
  author={Zheng, Zihua and Nie, Ni and Ling, Zhi and Xiong, Pengfei and Liu, Jiangyu and Wang, Hao and Li, Jiankun},
  booktitle=CVPR,
  pages={8925--8934},
  year={2022}
}

@inproceedings{menze2015object,
  title={Object scene flow for autonomous vehicles},
  author={Menze, Moritz and Geiger, Andreas},
  booktitle=CVPR,
  pages={3061--3070},
  year={2015}
}

@inproceedings{perazzi2016benchmark,
  title={A benchmark dataset and evaluation methodology for video object segmentation},
  author={Perazzi, Federico and Pont-Tuset, Jordi and McWilliams, Brian and Van Gool, Luc and Gross, Markus and Sorkine-Hornung, Alexander},
  booktitle = CVPR,
  pages={724--732},
  year={2016}
}

@inproceedings{he2016deep,
  title={Deep residual learning for image recognition},
  author={He, Kaiming and Zhang, Xiangyu and Ren, Shaoqing and Sun, Jian},
  booktitle = CVPR,
  pages={770--778},
  year={2016}
}

@inproceedings{liu2026arflow,
  title={Arflow: Auto-regressive optical flow estimation for arbitrary-length videos via progressive next-frame forecasting},
  author={Liu, Jiuming and Liu, Mengmeng and Zhu, Siting and Zhang, Yunpeng and Li, Jiangtao and Yang, Michael Ying and Nex, Francesco and Cheng, Hao and Wang, Hesheng},
  booktitle={The Fourteenth International Conference on Learning Representations},
  year={2026}
}

@article{aydemir2024can,
  title={Can visual foundation models achieve long-term point tracking?},
  author={Aydemir, G{\"o}rkay and Xie, Weidi and G{\"u}ney, Fatma},
  journal={arXiv preprint arXiv:2408.13575},
  year={2024}
}

@inproceedings{aydemir2025trackon,
  title={Track-On: Transformer-based Online Point Tracking with Memory},
  author={Aydemir, G{\"o}rkay and Cai, Xiongyi and Xie, Weidi and G{\"u}ney, Fatma},
  booktitle={The Thirteenth International Conference on Learning Representations},
  year={2025}
}

@article{aydemir2025trackon2,
  title={Track-On2: Enhancing Online Point Tracking with Memory},
  author={Aydemir, G{\"o}rkay and Xie, Weidi and G{\"u}ney, Fatma},
  journal={arXiv preprint arXiv:2509.19115},
  year={2025}
}

@article{sand2008particle,
  title={Particle video: Long-range motion estimation using point trajectories},
  author={Sand, Peter and Teller, Seth},
  journal={International journal of computer vision},
  volume={80},
  number={1},
  pages={72--91},
  year={2008},
  publisher={Springer}
}

@inproceedings{wang2025vggt,
  title={VGGT: Visual Geometry Grounded Transformer},
  author={Wang, Jianyuan and Chen, Minghao and Karaev, Nikita and Vedaldi, Andrea and Rupprecht, Christian and Novotny, David},
  booktitle={Proceedings of the IEEE/CVF Conference on Computer Vision and Pattern Recognition},
  year={2025}
}

@article{keetha2025mapanything,
  title={MapAnything: Universal feed-forward metric 3D reconstruction},
  author={Keetha, Nikhil and M{\"u}ller, Norman and Sch{\"o}nberger, Johannes and Porzi, Lorenzo and Zhang, Yuchen and Fischer, Tobias and Knapitsch, Arno and Zauss, Duncan and Weber, Ethan and Antunes, Nelson and others},
  journal={arXiv preprint arXiv:2509.13414},
  year={2025}
}

@article{zhang2025ufm,
  title={UFM: A Simple Path towards Unified Dense Correspondence with Flow},
  author={Zhang, Yuchen and Keetha, Nikhil and Lyu, Chenwei and Jhamb, Bhuvan and Chen, Yutian and Qiu, Yuheng and Karhade, Jay and Jha, Shreyas and Hu, Yaoyu and Ramanan, Deva and others},
  journal={arXiv preprint arXiv:2506.09278},
  year={2025}
}

@misc{wang2025pi3,
      title={$\pi^3$: Scalable Permutation-Equivariant Visual Geometry Learning}, 
      author={Yifan Wang and Jianjun Zhou and Haoyi Zhu and Wenzheng Chang and Yang Zhou and Zizun Li and Junyi Chen and Jiangmiao Pang and Chunhua Shen and Tong He},
      year={2025},
      eprint={2507.13347},
      archivePrefix={arXiv},
      primaryClass={cs.CV},
      url={https://arxiv.org/abs/2507.13347}, 
}

@article{lin2025movies,
  title={MoVieS: Motion-aware 4D dynamic view synthesis in one second},
  author={Lin, Chenguo and Lin, Yuchen and Pan, Panwang and Yu, Yifan and Yan, Honglei and Fragkiadaki, Katerina and Mu, Yadong},
  journal={arXiv preprint arXiv:2507.10065},
  year={2025}
}

@inproceedings{ma2022multiview,
  title={Multiview stereo with cascaded epipolar raft},
  author={Ma, Zeyu and Teed, Zachary and Deng, Jia},
  booktitle={European Conference on Computer Vision},
  pages={734--750},
  year={2022},
  organization={Springer}
}

@inproceedings{zhao2022global,
  title={Global matching with overlapping attention for optical flow estimation},
  author={Zhao, Shiyu and Zhao, Long and Zhang, Zhixing and Zhou, Enyu and Metaxas, Dimitris},
  booktitle={Proceedings of the IEEE/CVF Conference on Computer Vision and Pattern Recognition},
  pages={17592--17601},
  year={2022}
}

@inproceedings{truong2020glu,
  title={GLU-Net: Global-local universal network for dense flow and correspondences},
  author={Truong, Prune and Danelljan, Martin and Timofte, Radu},
  booktitle={Proceedings of the IEEE/CVF conference on computer vision and pattern recognition},
  pages={6258--6268},
  year={2020}
}

@inproceedings{luo2024flowdiffuser,
  title={Flowdiffuser: Advancing optical flow estimation with diffusion models},
  author={Luo, Ao and Li, Xin and Yang, Fan and Liu, Jiangyu and Fan, Haoqiang and Liu, Shuaicheng},
  booktitle={Proceedings of the IEEE/CVF Conference on Computer Vision and Pattern Recognition},
  pages={19167--19176},
  year={2024}
}

@misc{oquab2023dinov2,
  title={DINOv2: Learning Robust Visual Features without Supervision},
  author={Oquab, Maxime and Darcet, Timothée and Moutakanni, Theo and Vo, Huy V. and Szafraniec, Marc and Khalidov, Vasil and Fernandez, Pierre and Haziza, Daniel and Massa, Francisco and El-Nouby, Alaaeldin and Howes, Russell and Huang, Po-Yao and Xu, Hu and Sharma, Vasu and Li, Shang-Wen and Galuba, Wojciech and Rabbat, Mike and Assran, Mido and Ballas, Nicolas and Synnaeve, Gabriel and Misra, Ishan and Jegou, Herve and Mairal, Julien and Labatut, Patrick and Joulin, Armand and Bojanowski, Piotr},
  journal={arXiv:2304.07193},
  year={2023}
}

@misc{simeoni2025dinov3,
  title={{DINOv3}},
  author={Sim{\'e}oni, Oriane and Vo, Huy V. and Seitzer, Maximilian and Baldassarre, Federico and Oquab, Maxime and Jose, Cijo and Khalidov, Vasil and Szafraniec, Marc and Yi, Seungeun and Ramamonjisoa, Micha{\"e}l and Massa, Francisco and Haziza, Daniel and Wehrstedt, Luca and Wang, Jianyuan and Darcet, Timoth{\'e}e and Moutakanni, Th{\'e}o and Sentana, Leonel and Roberts, Claire and Vedaldi, Andrea and Tolan, Jamie and Brandt, John and Couprie, Camille and Mairal, Julien and J{\'e}gou, Herv{\'e} and Labatut, Patrick and Bojanowski, Piotr},
  year={2025},
  eprint={2508.10104},
  archivePrefix={arXiv},
  primaryClass={cs.CV},
  url={https://arxiv.org/abs/2508.10104},
}

@article{depth_anything_v2,
  title={Depth Anything V2},
  author={Yang, Lihe and Kang, Bingyi and Huang, Zilong and Zhao, Zhen and Xu, Xiaogang and Feng, Jiashi and Zhao, Hengshuang},
  journal={arXiv:2406.09414},
  year={2024}
}

@article{wen2025stereo,
  title={FoundationStereo: Zero-Shot Stereo Matching},
  author={Bowen Wen and Matthew Trepte and Joseph Aribido and Jan Kautz and Orazio Gallo and Stan Birchfield},
  journal={CVPR},
  year={2025}
}

@inproceedings{zhou2024samflow,
  title={Samflow: Eliminating any fragmentation in optical flow with segment anything model},
  author={Zhou, Shili and He, Ruian and Tan, Weimin and Yan, Bo},
  booktitle={Proceedings of the AAAI Conference on Artificial Intelligence},
  volume={38},
  pages={7695--7703},
  year={2024}
}

@InProceedings{flyingthings,
  author    = "N. Mayer and E. Ilg and P. H{\"a}usser and P. Fischer and D. Cremers and A. Dosovitskiy and T. Brox",
  title     = "A Large Dataset to Train Convolutional Networks for Disparity, Optical Flow, and Scene Flow Estimation",
  booktitle = "IEEE International Conference on Computer Vision and Pattern Recognition (CVPR)",
  year      = "2016",
  note      = "arXiv:1512.02134",
  url       = "http://lmb.informatik.uni-freiburg.de/Publications/2016/MIFDB16"
}

@inproceedings{tartanair2020iros,
  title={Tartanair: A dataset to push the limits of visual slam},
  author={Wang, Wenshan and Zhu, Delong and Wang, Xiangwei and Hu, Yaoyu and Qiu, Yuheng and Wang, Chen and Hu, Yafei and Kapoor, Ashish and Scherer, Sebastian},
  booktitle={2020 IEEE/RSJ International Conference on Intelligent Robots and Systems (IROS)},
  pages={4909--4916},
  year={2020},
  organization={IEEE}
}

@inproceedings{deng2009imagenet,
  title={Imagenet: A large-scale hierarchical image database},
  author={Deng, Jia and Dong, Wei and Socher, Richard and Li, Li-Jia and Li, Kai and Fei-Fei, Li},
  booktitle={2009 IEEE conference on computer vision and pattern recognition},
  pages={248--255},
  year={2009},
  organization={Ieee}
}

@inproceedings{liu2022convnet,
  title={A convnet for the 2020s},
  author={Liu, Zhuang and Mao, Hanzi and Wu, Chao-Yuan and Feichtenhofer, Christoph and Darrell, Trevor and Xie, Saining},
  booktitle={Proceedings of the IEEE/CVF conference on computer vision and pattern recognition},
  pages={11976--11986},
  year={2022}
}

@article{loshchilov2017decoupled,
  title={Decoupled weight decay regularization},
  author={Loshchilov, Ilya and Hutter, Frank},
  journal={arXiv preprint arXiv:1711.05101},
  year={2017}
}

@inproceedings{ranftl2021vision,
  title={Vision transformers for dense prediction},
  author={Ranftl, Ren{\'e} and Bochkovskiy, Alexey and Koltun, Vladlen},
  booktitle={Proceedings of the IEEE/CVF international conference on computer vision},
  pages={12179--12188},
  year={2021}
}

@inproceedings{shi2016real,
  title={Real-time single image and video super-resolution using an efficient sub-pixel convolutional neural network},
  author={Shi, Wenzhe and Caballero, Jose and Husz{\'a}r, Ferenc and Totz, Johannes and Aitken, Andrew P and Bishop, Rob and Rueckert, Daniel and Wang, Zehan},
  booktitle={Proceedings of the IEEE conference on computer vision and pattern recognition},
  pages={1874--1883},
  year={2016}
}

@article{shah2024flashattention,
  title={Flashattention-3: Fast and accurate attention with asynchrony and low-precision},
  author={Shah, Jay and Bikshandi, Ganesh and Zhang, Ying and Thakkar, Vijay and Ramani, Pradeep and Dao, Tri},
  journal={Advances in Neural Information Processing Systems},
  volume={37},
  pages={68658--68685},
  year={2024}
}

@inproceedings{jung2023anyflow,
  title={Anyflow: Arbitrary scale optical flow with implicit neural representation},
  author={Jung, Hyunyoung and Hui, Zhuo and Luo, Lei and Yang, Haitao and Liu, Feng and Yoo, Sungjoo and Ranjan, Rakesh and Demandolx, Denis},
  booktitle={Proceedings of the IEEE/CVF Conference on Computer Vision and Pattern Recognition},
  pages={5455--5465},
  year={2023}
}

@inproceedings{sui2022craft,
  title={Craft: Cross-attentional flow transformer for robust optical flow},
  author={Sui, Xiuchao and Li, Shaohua and Geng, Xue and Wu, Yan and Xu, Xinxing and Liu, Yong and Goh, Rick and Zhu, Hongyuan},
  booktitle={Proceedings of the IEEE/CVF conference on Computer Vision and Pattern Recognition},
  pages={17602--17611},
  year={2022}
}

@Article{paszke2019pytorchai,
  title        = "PyTorch: An Imperative Style, High-Performance Deep Learning Library",
  author       = "Adam Paszke and S. Gross and Francisco Massa and A. Lerer and J. Bradbury and G. Chanan and T. Killeen and Z. Lin and N. Gimelshein and L. Antiga and Alban Desmaison and Andreas K{\"o}pf and E. Yang and Zach DeVito and Martin Raison and Alykhan Tejani and Sasank Chilamkurthy and B. Steiner and Lu Fang and Junjie Bai and Soumith Chintala",
  journal      = "ArXiv",
  year         = "2019",
}

@article{doersch2022tap,
  title={Tap-vid: A benchmark for tracking any point in a video},
  author={Doersch, Carl and Gupta, Ankush and Markeeva, Larisa and Recasens, Adria and Smaira, Lucas and Aytar, Yusuf and Carreira, Joao and Zisserman, Andrew and Yang, Yi},
  journal={Advances in Neural Information Processing Systems},
  volume={35},
  pages={13610--13626},
  year={2022}
}

@inproceedings{vecerik2024robotap,
  title={Robotap: Tracking arbitrary points for few-shot visual imitation},
  author={Vecerik, Mel and Doersch, Carl and Yang, Yi and Davchev, Todor and Aytar, Yusuf and Zhou, Guangyao and Hadsell, Raia and Agapito, Lourdes and Scholz, Jon},
  booktitle={2024 IEEE International Conference on Robotics and Automation (ICRA)},
  pages={5397--5403},
  year={2024},
  organization={IEEE}
}

@inproceedings{greff2022kubric,
  title={Kubric: A scalable dataset generator},
  author={Greff, Klaus and Belletti, Francois and Beyer, Lucas and Doersch, Carl and Du, Yilun and Duckworth, Daniel and Fleet, David J and Gnanapragasam, Dan and Golemo, Florian and Herrmann, Charles and others},
  booktitle={Proceedings of the IEEE/CVF conference on computer vision and pattern recognition},
  pages={3749--3761},
  year={2022}
}

@inproceedings{smith2019super,
  title={Super-convergence: Very fast training of neural networks using large learning rates},
  author={Smith, Leslie N and Topin, Nicholay},
  booktitle={Artificial intelligence and machine learning for multi-domain operations applications},
  volume={11006},
  pages={369--386},
  year={2019},
  organization={SPIE}
}

@inproceedings{sun2010secrets,
  title={Secrets of optical flow estimation and their principles},
  author={Sun, Deqing and Roth, Stefan and Black, Michael J},
  booktitle={2010 IEEE computer society conference on computer vision and pattern recognition},
  pages={2432--2439},
  year={2010},
  organization={IEEE}
}
